\documentclass{article}

\usepackage{array}
\newcolumntype{L}[1]{>{\raggedright\arraybackslash}p{#1}}
\usepackage{adjustbox}
\usepackage[most]{tcolorbox}
\usepackage{pifont}
\usepackage{enumitem}
\usepackage{parskip}
\usepackage{hyperref}
\usepackage{float}
\usepackage{longtable}
\usepackage{url}
\usepackage{subcaption}
\usepackage[table]{xcolor}
\usepackage{multirow}

\definecolor{expBlue}{HTML}{1F4E79}
\definecolor{expTeal}{HTML}{146C6C}
\definecolor{expPlum}{HTML}{6A1B7A}
\definecolor{softBg}{HTML}{F7F9FC}
\usepackage{pdflscape}   
\definecolor{softgreen}{RGB}{70,130,80}
\definecolor{softorange}{RGB}{195,120,40}
\definecolor{softred}{RGB}{175,60,60}
\newtcolorbox{promptbox}[2][]{
  enhanced,
  breakable,
  colback=softBg,
  colframe=#2,
  coltitle=white,
  fonttitle=\bfseries,
  title=#1,
  attach boxed title to top left={xshift=6pt,yshift=-\tcboxedtitleheight/2},
  boxed title style={colback=#2, sharp corners, boxrule=0pt},
  arc=3pt,
  boxrule=0.6pt,
  left=10pt, right=10pt, top=12pt, bottom=10pt,
  before skip=14pt, after skip=14pt,
}
\urldef{\BAOorganismURL}\url{https://bioportal.bioontology.org/ontologies/BAO?p=classes&conceptid=http%3A%2F%2Fwww.bioassayontology.org%2Fbao%23BAO_0000218}

\usepackage[dblblindworkshop, final, nonatbib]{neurips_2026}

\usepackage[utf8]{inputenc} 
\usepackage[T1]{fontenc}    
\usepackage{booktabs}       
\usepackage{amsfonts}       
\usepackage{nicefrac}       
\usepackage{microtype}      
\definecolor{grey}{rgb}{0.5,0.5,0.5}
\usepackage{csquotes}  
\usepackage[
  backend=biber,
  style=numeric-comp,   
  sorting=none,         
  maxbibnames=3,        
  minbibnames=1,        
  maxcitenames=5,       
  mincitenames=1,
  giveninits=true,      
  natbib=true,          
  uniquelist=false,
  uniquename=false,
  doi=true, url=false, isbn=false, eprint=false
]{biblatex}
\addbibresource{references.bib}
\DeclareSourcemap{
  \maps[datatype=bibtex]{
    \map{
      \step[fieldset=abstract,  null]
      \step[fieldset=pdf,       null]
      \step[fieldset=copyright, null]
      \step[fieldset=keywords,  null]
      \step[fieldset=language,  null]
    }
  }
}

\title{Can LLMs Reliably Annotate Bioassay Metadata to Improve Data Readiness?}
\workshoptitle{AI Data Readiness for Scientific Discovery}

\usepackage{graphicx}

\author{
\textbf{Laura van Weesep\textsuperscript{*,1,2}},
\textbf{Riccardo Tedoldi\textsuperscript{*,1,3}},
\textbf{Jens Sjölund\textsuperscript{2}},
\textbf{Hossein Azizpour\textsuperscript{3,4}},\\
\textbf{Susanne Winiwarter\textsuperscript{1,5}},
\textbf{Ola Engkvist\textsuperscript{1,6}},
\textbf{Jon Paul Janet\textsuperscript{1}},\\
\textbf{Samuel Genheden\textsuperscript{1}},
\textbf{Juan Viguera Diez\textsuperscript{1}}
\\[8pt]
 \textsuperscript{1}Molecular AI, Discovery Sciences R\&D, AstraZeneca\\
 \textsuperscript{2}Department of Information Technology, Uppsala University\\
 \textsuperscript{3}Robotics, Perception \& Learning, KTH Royal Institute of Technology\\
\textsuperscript{4}Science for Life Laboratory, Stockholm, Sweden\\
\textsuperscript{5}Drug Metabolism and Pharmacokinetics, Research and Early Development, 
Cardiovascular,\\ Renal and Metabolism (CVRM), BioPharmaceuticals R\&D, AstraZeneca\\
\textsuperscript{6}Department of Computer Science and Engineering,\\
Chalmers University of Technology and University of Gothenburg, Sweden
\\[8pt]
\textsuperscript{*}equal contributors\\
 \small{
   \textbf{Correspondence:}  \href{mailto:lauradesire.vanweesep@astrazeneca.com}{lauradesire.vanweesep@astrazeneca.com},
   \href{mailto:riccardo.tedoldi@astrazeneca.com}{riccardo.tedoldi@astrazeneca.com}
 }
}

\usepackage{url}

\begin{document}

\maketitle

\begin{abstract}
The emergence of foundation models for molecular property prediction requires a high degree of AI data readiness,  including reliable metadata annotation. However, both public repositories and industrial screening databases suffer from missing, inconsistent, or conflated assay annotations. In this work, we quantify the extent of missing annotations in PubChem for the BioAssay Ontology (BAO) assay format and physical detection method fields and investigate whether open-source and proprietary large language models (LLMs) can reliably predict and audit metadata annotations directly from the assay text.
In our assessment, we found that the annotation coverage across PubChem's $\sim$2 million bioassays is critically sparse, 36\% lacking an assay format, 89\% a BioAssay type, and >99.9\% any BAO-mapped assay format or detection technology term. This motivates the need for automated test-metadata curation. Using evaluation sets derived from PubChem and ChEMBL, we assess the agreement of seven open-source and proprietary LLMs with existing silver labels. Recall is at least 0.96 for biochemical and cell-based assay formats, with a similar pattern for detection technology, although disagreements increase on under-represented classes. Manual inspection shows that many of these disagreements trace back to inconsistencies between silver sources rather than to LLM error. Moreover, in a qualitative study with a senior industrial curator, LLM-generated evidence prompted the expert to revise some of  their own labels, showing LLMs can flag potentially mislabeled assays. Across the study, performance differences between proprietary and open-source models were small. Together, these results suggest LLMs can support the large-scale annotation and auditing of assay metadata, though per-class reliability estimates and targeted human review remain necessary before such labels enter downstream ML pipelines.

\end{abstract}

\section{Introduction}
\label{sec:intro}
The FAIR data principles~\cite{wilkinson2016fair} introduced standards to make data findable, accessible, interoperable, and reusable, yet AI readiness for many datasets remains limited. In particular, missing, inconsistent, and incorrect metadata remain widespread problems in scientific data repositories.
Bioassay collections are a notable example and a fundamental source of data for machine learning models for drug discovery. 

A bioassay measures the effect of a substance on a biological system to obtain insight into the expected biological activity. PubChem~\cite{Kim2024} is the largest public chemical database, containing more than 100 million compounds and bioassay results from more than a million assays. Many entries lack structured metadata on the nature of the experiment, such as the assay format and the detection method. The lack of metadata in PubChem reflects issues such as records that predate relevant standards or inconsistent use of terminology among contributors. Industrial assay databases present related problems, including missing historical data and the use of single labels, such as the ``assay category'' to describe several distinct aspects of an assay at once.

The BioAssay Ontology (BAO)~\cite{Visser2011, Abeyruwan2014} addresses the ambiguity challenge by decomposing assays into orthogonal axes, including the biological format and the physical detection method, the endpoint measured and the target with each a formal class definition. This distinction allows for assays of the same format but with a different detection method to be distinguished. It aims to provide a framework for mapping unstructured descriptions to standardized entities. PubChem's native ``BioAssay Type'' field mixes format, modality, and endpoint into composite labels such as In vivo|Toxicity or Cell-based|RNAi for instance. Disentangling such labels retroactively or filling in missing labels manually does not scale to the level of millions of PubChem records or decades of legacy industrial assays.

Large language models offer a promising approach to close this gap. For instance, instruction-tuned LLMs have recently been shown to perform well on ontology alignment in biomedical fields \cite{Cinquin2024,Ikeda2025}. Nevertheless, their performance on ontologies with multiple dimensions, such as BAO, remains underexplored. We address this gap by evaluating LLMs on the assignment of BAO assay formats and assay detection methods. 

Our contributions can be summarized as: (i) a systematic analysis of missing and inconsistent assay format and detection technique metadata in PubChem BioAssay's bulk-download distribution, (ii) an assessment of four open-source and three closed-source LLMs on text-conditioned BAO assay format classification, using ChEMBL cross-references as a reference label followed by further exploration for physical detection method and subsets of PubChem and (iii) manual review of LLM vs. current-label disagreement, further backed up by a curator, in which we found examples of a senior industrial curator agreeing with the LLM over the original dataset label. Together, these results show that LLMs can detect likely mislabeled assays and in the future support automated annotation\footnote{Code and data will be publicly released upon acceptance of the full paper.}.

\section{Background and Related Work}
\label{sec:related}

\subsection{Bioassay Ontology (BAO)}
{The BioAssay Ontology (BAO)} was introduced to give the output of high-throughput screening campaigns explicit description-logic semantics rather than free-text labels~\citep{Visser2011}, and later extended and modernized to cover a wider range of assay types and to interoperate with adjacent ontologies (cell line, target, and unit vocabularies)~\citep{Abeyruwan2014}. The BAO model focuses on subdividing assays into orthogonal axes, including categories related to the bioassay component, biology, endpoint, format, and method component. However, achieving comprehensive coverage of all bioassays has proved challenging. For example, in 2014, BAO could not adequately represent all transporter assays in ChEMBL~\citep{zdrazil2014transporter}. Since then, it has undergone numerous updates to expand its capacity to describe and categorize diverse assay types. The current version (uploaded on the 6th of May 2026 \texttt{v2.8.19}) contains 7918 separate classes\footnote{\url{http://www.bioassayontology.org/}} and is a product of over a decade of iterative validation against real-world chemical biology datasets.

Early efforts to apply BAO relied on manual annotation. \citet{vempati2012formalization} manually annotated 944 PubChem assays using a drop-down menu. They showed that they could map 299 distinct PubChem names to 20 controlled BAO terms, demonstrating the potential of the ontology to reduce terminological ambiguity and reveal additional relationships between assays. Industry has also adopted the BAO framework. AstraZeneca utilized this framework to annotate 381 internal assays run from 2005--2013, along with 233 PubChem assays to get insight into the technology gaps between their internal screening and external screening, compare compound–activity overlap and flag technology-specific artifacts~\cite{zander2015using}. However, manual annotation of protocols remains laborious and does not scale to the millions of bioassays available in public data repositories.

\subsection{Annotated public data repositories} 
In this work, we use three sources of bioassay annotations: PubChem, ChEMBL and BARD. PubChem~\cite{Kim2024} is the most extensive collection of publicly available data from high-throughput screening and medicinal chemistry experiments, which contains descriptions of protocols, target and results that many screening laboratories, biotech, pharmaceutical companies, and academic institutions have deposited since 2004~\cite{Wang2011}, currently comprising $\sim$1.9M bioassays (AIDs) as of the 2026 bulk download. However, records vary greatly on how and if the experimental metadata was reported, and often the information is present in form of unstructured free-text. 

To address this, the BioAssay Research Database (BARD)~\cite{howe2015bioassay} produced a curated set of BAO annotations on PubChem records from the NIH Molecular Libraries Program (MLP). Curators mapped free-text PubChem protocols onto BAO terms, and the annotations were reviewed by MLP scientists involved in the original experiments. Although the BARD portal has been retired, its 4 477 curated annotations for 1 000 assays remain available in PubChem~\cite{ncbi_bard} as a bulk download. Out of these assays 997 have an annotated detection method and 998 had an annotated assay format. The BARD annotations for these assays and some assays beyond this, appear under the ``BioAssay Annotations''. 

ChEMBL, in turn, is a manually curated bioactivity database that collects bioactivity data extracted primarily from the scientific literature~\cite{Gaulton2011, Mendez2019-zf} and based on voluntary data depositors comprising $\sim$2.0M assays. ChEMBL curators assign standardized assay and target metadata and link some records to PubChem assay ID (AID) identifiers, an independent source of reference annotations for a subset of PubChem assays. ChEMBL has also introduced direct-deposition workflows with structured submission formats, depositor-defined identifiers, and validation procedures, aimed at improving metadata consistency ~\cite{Mendez2019-zf, Zdrazil2023}. Nevertheless, many of these annotations lack meaningful BAO terms; $\sim$10\% of the assays have unspecified assay formats, detection-method labels are entirely absent, and some existing annotations were generated using rule-based methods that may be unreliable~\cite{smit2026enhancing}.

These three annotation sources differ not only in their coverage of assays and metadata coverage but also in the way their labels are produced. Native PubChem metadata are derived from voluntary, free-text depositor submissions and have been found to lack the consistent structure needed for cross-assay comparison \cite{howe2015bioassay}. ChEMBL improves on this through expert bioactivity curation, although the metadata coverage is limited. BARD was purpose-built for accuracy, and its expert validation step is absent from both PubChem's native fields and ChEMBL's assay annotations to our knowledge. However it only covers a small subset of PubChem.  For assays annotated in all three sources, we therefore treat BARD-derived labels as the most trustworthy reference, ChEMBL as a secondary also partial source, and native PubChem metadata as the least reliable baseline.

Because of these reliability gaps, several groups have built curated subsets or rule-based cleanup pipelines rather than fixing annotations at the source. Curated benchmarks such as LIT-PCBA~\cite{Tran-Nguyen2020-vu} and MF-PCBA~\cite{Buterez2023} clean HTS data and discard unreliable records, while ScrubChem applies logic-based text-processing rules to curate PubChem~\cite{harris2017scrubchem}. More recently, CAPRICHO~\cite{Araripe2026} addressed curation of ChEMBL bioactivity data for QSAR modeling by flagging quality issues rather than removing them, letting users assess how each curation decision impacts the results.
Its case studies show that assay metadata strongly affects whether measurements can be combined: grouping by assay type and cell type improved cross-assay agreement. 
However, CAPRICHO relies on existing ChEMBL metadata fields and on keyword-based queries of free-text assay descriptions. This highlights the need for more complete and structured assay annotations, which we address here using LLMs.

Overall, although public repositories contain millions of bioassays, their BAO annotations are often missing, incomplete or unreliable, motivating the development of scalable, automated annotation methods.

\subsection{Bioassay curation with language models} 
LLMs offer a promising approach to automated assay annotation. LLMs are increasingly used to extract metadata from biomedical records. For example, ChIP-GPT uses a LLaMA-based model to extract chromatin immunoprecipitation metadata from free-text records~\cite{Cinquin2024}. Similarly, ~\citet{Ikeda2025} used Llama 3 to extract cell line and gene perturbation terms from BioSample descriptions. In both studies, the LLM method outperformed the traditional mapping approach. For assay format annotation specifically, recent studies have explored AI-assisted pipelines. One approach uses a named entity recognition (NER) model to extract the assay type from a curated description, followed by text2term~\cite{gonccalves2024text2term} to ground free-text metadata to biomedical ontology terms~\cite{smit2026enhancing}. Other methods, including a retrieval-then-prompt framework~\cite{Taboada2025, Liu2021} and other similar pipelines such as RELATE~\cite{pmlr-v297-olasunkanmi26a}, ground LLM outputs to formal ontology terms rather than free text. This helps address a common issue with LLMs: their tendency to generate non-standard labels. To our knowledge, whether the same strategies could reliably assign BAO labels (i.e., format, detection, target) from free text consistently is unexplored and we aim to address this gap.

\section{Methods}
\label{sec:methods}

\paragraph{Dataset Construction.}  To evaluate the performance of LLMs for predicting bioassay metadata classes, we constructed our dataset from the PubChem BioAssay bulk JSON data and combined this with the associated annotations of assays and comment files (detailed in Appendix~\ref{app:dataset-creation}).  This process yielded
1 770 568 AIDs (containing a name, comment, and/or description) out
of 1 994 310 downloaded in total.  Because some metadata fields were empty or missing in the bulk files despite being available through the PubChem web interface, we additionally queried PubChem's PUG REST endpoint for the assays included in our analysis. We used the returned records to supplement the assay title, description, comments and protocol when available. 

We also queried the ChEMBL 37 database\footnote{\url{https://ftp.ebi.ac.uk/pub/databases/ChEMBL/ChEMBLdb/releases/ChEMBL_37/ChEMBL_37_sqlite.tar.gz}} and retained records that could be matched to a PubChem assay. We extracted the corresponding BAO assay format annotations, excluded PubChem AIDs that mapped to multiple ChEMBL entries, and discarded entries whose BAO assay format label was \texttt{assay format}. We also removed records with duplicate descriptions to not feed the same text field to the LLM multiple times. We followed the same de-duplication procedure and label cleanup for the AIDs for which BAO assay annotations were available in PubChem but not in ChEMBL. We similarly extracted and filtered data from assays containing physical detection method metadata.

\begin{figure}[!h]
    \noindent\makebox[\textwidth]{%
        \includegraphics[width=1.1\textwidth]{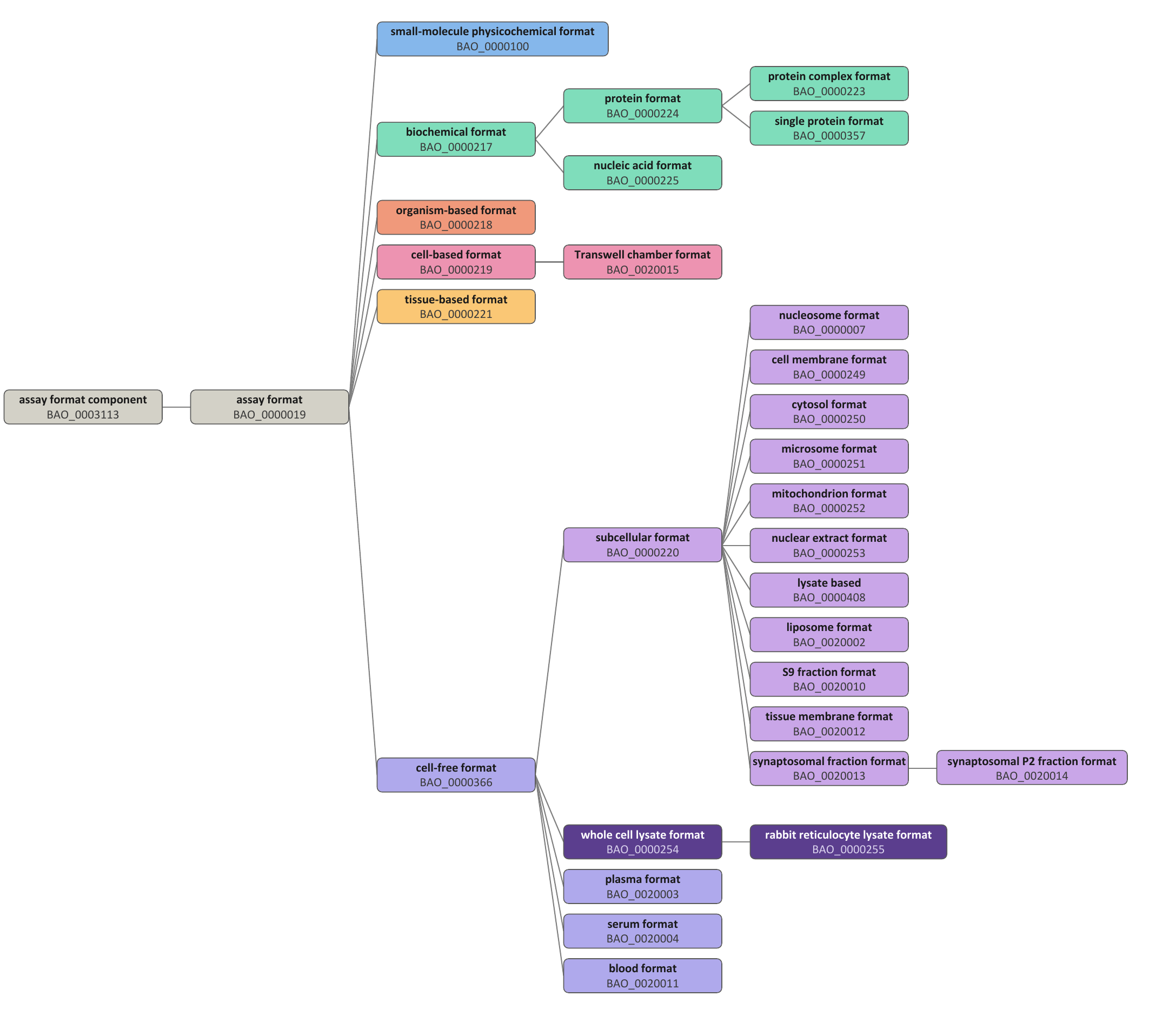}%
    }
    \caption{Figure was made by downloading the BAO complete OWL and creating the subtree in PowerPoint.}
    \label{fig:tree_assay_format}
\end{figure}

\begin{figure}[!h]
    \noindent\makebox[\textwidth]{%
        \includegraphics[width=1.4\textwidth]{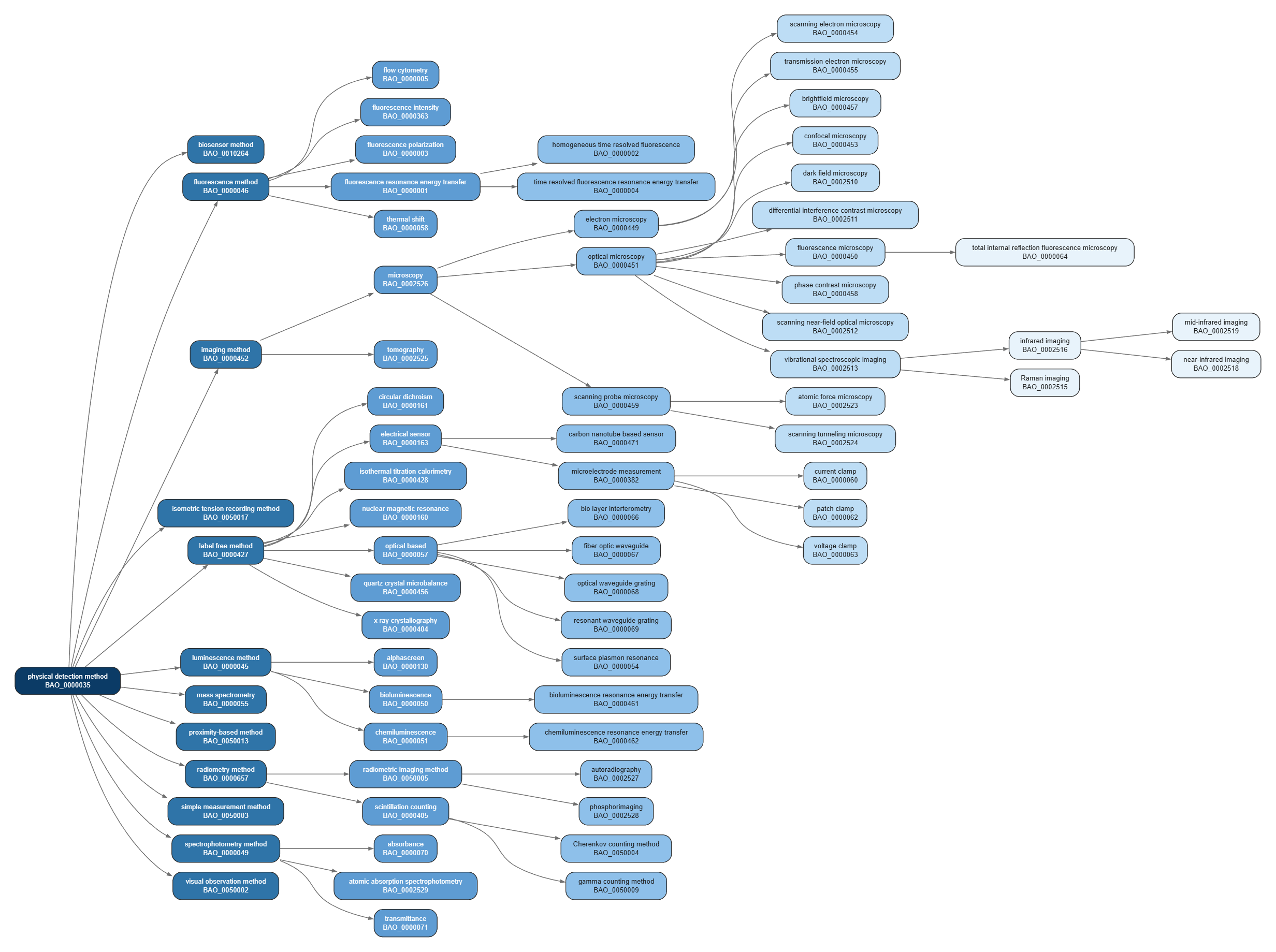}%
    }
    \caption{Figure was made by downloading the BAO complete OWL and creating the subtree in PowerPoint.}
    \label{fig:bao_assay_detection_tree}
\end{figure}

We mapped assay format and physical detection method labels to depth-one categories in the BAO hierarchy (Appendix Fig.~\ref{fig:tree_assay_format} and Appendix Fig.~\ref{fig:bao_assay_detection_tree}). PubChem and ChEMBL labels that could not be mapped to a depth-one BAO category were discarded. This procedure yielded three final evaluation sets: 1 097 PubChem assays with assay format annotations from ChEMBL, 226 PubChem-only annotated assays with assay format labels, and 308 PubChem assays with physical detection method labels.

Because the available reference labels are inherited from existing database annotations that included inconsistencies and disagreements with current annotation standards, we refer to them as ``silver labels'' as they might not reflect the ground truth according to the current BAO system. In this study, silver labels are derived from existing PubChem, ChEMBL, and BioAssay Research Database annotations and are used to evaluate agreement with LLM-generated labels.

\paragraph{Task definition and LLMs utilized.}
We prompt LLMs to predict the assay format or assay physical detection method based on the title, description, comments, and protocol.
We used three proprietary models: GPT-4o, Gemini 3.6 Flash, and Claude Sonnet 4.6, all accessed through APIs. We also evaluated four open-weights models: Gemma 4 31B, GPT-OSS 20B, Qwen 3.5, and Llama 3.3 70B, run locally via \texttt{Ollama v0.17.4}. Further details on the parameters used to run the models can be found in Appendix \ref{app: model_params}. For prompt-ablation experiments, we selected a stratified subset of 500 assays from the ChEMBL assay format set and evaluated the same assays under different prompt conditions, including alternative label orderings and prompts without the definition of the label. The complete prompts can be found in Appendix \ref{app: prompts_used}. For models run through Ollama, we enforced the label vocabulary at generation time using Ollama's JSON-Schema \texttt{format=} parameter.

\section{Missingness and Inconsistency in PubChem Bioassay Metadata}\label{sec: Missingness_inconsistencies}

We first investigated the quality and completeness of the metadata provided in the PubChem BioAssay bulk distribution. Out of the 1 994 310 assays in the bulk downloaded, only 89\% had a name, description, and comment, with less than 10\% unique descriptions. We followed up by analyzing the metadata (see Appendix Table~\ref{tab:assay_fields}), focusing specifically on four metadata categories: \textit{Assay Format}, \textit{BAO Assay Format}, \textit{BAO Detection Technology}, and \textit{BioAssay Type}. 
In Appendix~\ref{app:dataset-creation}, we report statistics of the distributions across the four metadata dimensions in Tables~\ref{tab:assay_formats},~\ref{tab:bao_assay_formats},~\ref{tab:bao_detection_technologies} and~\ref{tab:bioassay_types}. The labels show substantial heterogeneity, both in their level of detail and in the aspect of the assay they describe: some refer to the assay modality, others to the endpoint measured. We initially examined the BioAssay Type field to see whether it could inform BAO assay format assignment, but found it unsuitable as the vocabulary conflates format (e.g., Cell-based, Biochemical), methodology (RNAi), and endpoint (Toxicity), and a non-negligible fraction of entries combine multiple labels that are inconsistent with the BAO ontology (Table~\ref{tab:bioassay_types}).

A related source of variation is the PubChem Assay Format field itself, which offers only three categories (Cell-based, Organism-based, Biochemical), whereas the BAO assay format branch is far richer. This limited vocabulary likely contributes to annotation discrepancies. When curators are forced to pick from three options they may assign a different label than they would given the full BAO hierarchy. This motivates the use of BAO labels for assessing whether LLMs can assign assay formats, instead of using the PubChem Assay Format field itself. We also noticed inconsistencies from the available data in the PubChem web interface. To illustrate, the AID 1885, is reported as \textit{Cell-based} in the web interface, but as \textit{Cell-based|Organism-based} in the bulk download. Moreover, inconsistencies can occur within a single PubChem entry: the same assay may be associated with multiple conflicting BAO labels. From the 236 assays that had both the BARD and ChEMBL assay format labels on PubChem in the bulk BARD download, for 24 assays there was a disagreement (10.2\%). Further details and case studies can be found in Appendix \ref{app:annotation-inconsistencies}. Additionally, the coverage of the metadata fields is limited. The data from the PubChem bulk download present severe amounts of missing data, with 36\% out of the 1.9 million entries missing the assay format, and less than 0.1\% of the assays having a BAO assay format and physical detection method annotated in the comments.

All in all, PubChem contains substantial bioassay annotation issues, including missing and inconsistent information.

\section{Results: An LLM-based Auditing Workflow}
\label{sec:experiments}
Given the missingness, heterogeneity, and inconsistencies identified in Section~\ref{sec: Missingness_inconsistencies}, we evaluated whether LLMs could provide a scalable approach to reviewing and assigning bioassay metadata labels. Model predictions were compared against existing annotations derived from PubChem, ChEMBL, and BARD, which we refer to as silver labels. 
Because these references may themselves be incomplete, inconsistent, or erroneous, the per-class classification metrics reported below should not be interpreted as definitive annotation accuracy. To mitigate these limitations, we manually inspect a subset, keeping in mind that annotations from different sources might differ in trustworthiness.

\subsection{Reliability against silver sources}
\label{sec:reliability}
We first assess agreement among the LLMs using Cohen's $\kappa$~\citep{cohen1960coefficient}. On the ChEMBL assay format evaluation set, all seven LLMs achieve pairwise Cohen's $\kappa > 0.9$, with similar agreement on the BAO detection tasks. The agreement drops on the smaller PubChem-BAO subset, but is still high with $\kappa > 0.8$ for all models except Gemma 3 27B, where we see lower agreement with the other LLMs. Full agreement matrices for all three experiments along with a definition of the Cohen's $\kappa$ are provided in Appendix~\ref{app:cohen}.

Next, we evaluate whether the consensus signal tracks silver-label sources, using ChEMBL assay format labels as a reference. Table \ref{tab:per_class_chembl} reports classification metrics for each assay format class. A majority voting consensus of LLMs agrees with the ChEMBL label in 96$\%$ of the cases. Cell-based and biochemical format are classified consistently well with a recall at least 96\% for all models including the open-source ones. Performance on the organism-based format varies more across models: Gemma 3 27B achieves the highest precision, but makes the fewest predictions for this class. The majority vote achieves a recall of 87\% on the organism-based format with a precision of 77\%, noticeably lower than the performance on the cell-based and biochemical format. The cell-free and tissue-based formats show substantially lower precision and recall for all models, with at most one correctly predicted instance in either class. These results might suggest that LLMs are unreliable for rare assay format classes (only 6 and 2 entries in this dataset). However, 
manual inspection showed that for 7 out of 8 of these assays the BARD label was actually different from the ChEMBL label and LLMs agreed with the BARD label and for the last, the expert curator agreed with the LLM prediction. We will elaborate on this in the following paragraph.

\begin{table}[t]
\centering
\caption{Per-class recall / precision on the ChEMBL baseline (assay format). Values shown as
R/P; support ($n_\text{true}$) shown once per column in the header row.
The two lowest-support classes
(\textcolor{grey}{\textbf{tissue-based}} and
\textcolor{grey}{\textbf{cell-free}}, each n$<$10) are too small for reliable aggregate metrics. Invalid predictions were excluded for these metrics. Precision is reported as “–” when no predictions were made.}
\label{tab:per_class_chembl}
\begin{adjustbox}{width=\textwidth}
\begin{tabular}{@{}l cc cc c c c c@{}}
\toprule
\textbf{Model} & \textbf{Acc.} & \textbf{mF1} & \textbf{mF1\textsubscript{top3}} & \textbf{cell-based} & \textbf{biochemical} & \textbf{organism-based} & \textcolor{grey}{\textbf{tissue-based}} & \textcolor{grey}{\textbf{cell-free}} \\
 & & & & (n=784) & (n=252) & (n=53) & \textcolor{grey}{(n=6)} & \textcolor{grey}{(n=2)} \\
\midrule
Claude Sonnet 4.6 & 0.963 & 0.560 & 0.934 & 0.98 / 0.99 & 0.96 / 0.98 & 0.88 / 0.83 & \textcolor{grey}{0.00 / --} & \textcolor{grey}{0.00 / 0.00} \\
Gemini 3.6 Flash  & 0.957 & 0.578 & 0.919 & 0.97 / 0.99 & 0.96 / 0.97 & 0.94 / 0.71 & \textcolor{grey}{0.00 / --} & \textcolor{grey}{0.50 / 0.08} \\
GPT-4o            & 0.954 & 0.551 & 0.919 & 0.96 / 0.99 & 0.96 / 0.97 & 0.91 / 0.74 & \textcolor{grey}{0.00 / --} & \textcolor{grey}{0.00 / 0.00} \\
\midrule
GPT-OSS 20B       & 0.956 & 0.602 & 0.908 & 0.97 / 0.98 & 0.96 / 0.97 & 0.77 / 0.79 & \textcolor{grey}{0.17 / 1.00} & \textcolor{grey}{0.00 / 0.00} \\
Gemma 3 27B       & 0.979 & 0.576 & 0.960 & 0.99 / 0.98 & 0.99 / 0.97 & 0.85 / 0.98 & \textcolor{grey}{0.00 / --} & \textcolor{grey}{0.00 / 0.00} \\
Gemma 4 31B       & 0.951 & 0.545 & 0.909 & 0.96 / 0.99 & 0.96 / 0.97 & 0.91 / 0.70 & \textcolor{grey}{0.00 / --} & \textcolor{grey}{0.00 / 0.00} \\
Llama 3.3 70B     & 0.976 & 0.572 & 0.954 & 0.99 / 0.98 & 0.99 / 0.98 & 0.85 / 0.94 & \textcolor{grey}{0.00 / --} & \textcolor{grey}{0.00 / 0.00} \\
\midrule
{Majority vote} & 0.961 & {0.553} & {0.921} & {0.97 / 0.98} & {0.97 / 0.97} & {0.87 / 0.77} & \textcolor{grey}{{0.00 / --}} & \textcolor{grey}{{0.00 / 0.00}} \\
\bottomrule
\end{tabular}
\end{adjustbox}
\end{table}

We further inspected the 43 assays in which the majority of models disagreed with the ChEMBL label in Appendix~\ref{app:full_disagreement_table}. For 29 of these assays, PubChem BARD labels are also available. As we show in Table~\ref{tab:disagreement_summary}, out of these, in 17 cases (59 $\%$) the LLM consensus agreed with the BARD PubChem label but not ChEMBL. This finding is particularly notable because BARD annotations are generated through a comparatively thorough process that includes additional verification not applied to the ChEMBL labels. For the cases in which ChEMBL and BARD agreed, but LLMs did not, further inspection showed that the models did not unanimously disagree on the ChEMBL label (for full details see Appendix Table~\ref{tab:Wrong_ChEMBL_Label}). If the ChEMBL label is correct in these instances, a structured debate among models could potentially help resolve the disagreement. 

\begin{table}[t]
\centering
\small
\caption{Categorization of the 43 ChEMBL assays where the majority of
LLMs disagreed with the ChEMBL BAO assay format label, and how many of
each pattern were sent for senior industrial curator review
(Table~\ref{tab:Expert_Review}).  }
\label{tab:disagreement_summary}
\begin{tabular}{@{}L{6.5cm} c c c@{}}
\toprule
\textbf{Disagreement pattern} & \textbf{Count} & \textbf{\%} & \textbf{Sent for review} \\
\midrule
LLMs support the BARD label against ChEMBL & 17 & 40\% & 4 \\
ChEMBL and BARD agree and LLMs disagree with both & 10 & 23\% & 10 \\
ChEMBL and BARD disagree and LLMs disagree with both (novel prediction) & 2 & 5\% & 2 \\
No BARD label available; ChEMBL-only ground truth & 14 & 33\% & 1 \\
\midrule
\textbf{Total} & \textbf{43} & \textbf{100\%} & \textbf{17} \\
\bottomrule
\end{tabular}
\end{table}

\paragraph{Expert Annotator Review}
For 17 assay AIDs, we conducted a further manual review by asking domain experts to classify each assay. We included all assays for which the ChEMBL and PubChem BARD labels agreed but the LLM consensus differed. In addition, we randomly selected four cases in which the LLM consensus agreed with the PubChem BARD label but disagreed with the ChEMBL label. We also included cases in which the LLM consensus disagreed with both the ChEMBL and BARD labels.

We shared with the expert annotator the same information as with the LLMs: title, description, comment and protocol. Experts highlighted that classifying these examples was challenging. In 6 cases they agreed with the majority of LLMs immediately. In further six cases, they agreed after seeing the evidence provided by the LLMs.  In four additional cases, the experts considered the LLM classification plausible—three of them after reviewing the model-provided evidence—but did not assign a definitive label. They noted that some assays could reasonably fit multiple classes or lacked sufficient information for an unambiguous classification. These observations suggest that uncertainty or multiple applicable labels may be worth representing explicitly in the database. 

We realized that the annotator initially placed substantial weight on the format type stated in the assay title, for which in many cases they changed their mind after seeing LLM evidence. For example, for AID 2695, the title was “Fluorescence Cell-Based Dose Screen to Determine Inhibitors of C. albicans Viability”. The expert prediction was cell-based and the LLM predicted organism format.
Upon closer inspection, and after reviewing the LLMs’ reasoning, the expert concluded that the LLM classification was correct. The expert had initially overlooked the fact that, according to the official BAO definition, “single-celled organisms” should be classified as an organism format. This could be a result of the discrepancy between the assay format and the BAO assay format on ChEMBL. Finally, for assay AID 588769, the LLMs identified a likely mislabeled datapoint in PubChem and agreed with the expert’s classification. For AID 1470, the expert considered the classification highly uncertain; nevertheless, the expert’s assessment was consistent with the LLM consensus and differed from both the ChEMBL and PubChem labels, providing further evidence of alignment between the LLMs and human judgement.

Together, these cases suggest that, when grounded in BAO definitions, LLM predictions may help identify errors in both database annotations and initial expert assessments. However, this capability was not consistent across all cases. For AID 504631, the expert agreed with the ChEMBL label rather than with the label supported by the LLM consensus. The expert correctly noted that the assay used a living organism, whereas the LLMs frequently confused this category with a cell-line format (see Appendix~\ref{app:full_disagreement_table}). Thus, although the LLMs identified the relevant evidence, they drew the wrong conclusion from it.

To follow up on this, we conducted a second experiment where we used the third-party BAO annotations already available in PubChem. In Table~\ref{tab:per_class_pubchem_vs_bard} we observe again  LLM agreement with BARD labels compared to third party PubChem annotations. In the majority of cases where the BARD label did not align with the third party BAO assay format, LLMs agree with BARD. This highlights that there is a lack of consensus on what the ground truth should be.

\begin{table}[ht]
\centering
\caption{Per-class recall / precision on the PubChem subset.
Rare classes in \textcolor{grey}{grey}. Tissue-based format has $n=0$ in the PubChem subset and is omitted.}
\label{tab:per_class_pubchem_vs_bard}
\begin{adjustbox}{width=\textwidth}
\begin{tabular}{@{}l cc cc c c c c@{}}
\toprule
& & \textbf{Acc.} & \textbf{mF1} & \textbf{mF1\textsubscript{top3}} & \textbf{biochemical} & \textbf{cell-based} & \textcolor{grey}{\textbf{organism-based}} & \textcolor{grey}{\textbf{cell-free}} \\
\midrule
\multicolumn{2}{@{}l}{\textit{vs.\ third party PubChem}} & & & & (n=120) & (n=100) & \textcolor{grey}{(n=4)} & \textcolor{grey}{(n=2)} \\
\midrule
Claude Sonnet 4.6 & R / P & 0.649 & 0.350 & 0.534 & 0.54 / 0.88 & 0.78 / 0.84 & \textcolor{grey}{0.50 / 0.07} & \textcolor{grey}{1.00 / 0.08} \\
Gemini 3.6 Flash  & R / P & 0.655 & 0.349 & 0.533 & 0.56 / 0.87 & 0.77 / 0.82 & \textcolor{grey}{0.50 / 0.07} & \textcolor{grey}{1.00 / 0.08} \\
GPT-4o            & R / P & 0.668 & 0.356 & 0.551 & 0.59 / 0.87 & 0.76 / 0.84 & \textcolor{grey}{0.50 / 0.09} & \textcolor{grey}{1.00 / 0.07} \\
\midrule
GPT-OSS 20B       & R / P & 0.650 & 0.353 & 0.549 & 0.53 / 0.89 & 0.79 / 0.84 & \textcolor{grey}{0.50 / 0.10} & \textcolor{grey}{1.00 / 0.06} \\
Gemma 3 27B       & R / P & 0.765 & 0.397 & 0.614 & 0.73 / 0.86 & 0.82 / 0.84 & \textcolor{grey}{0.50 / 0.14} & \textcolor{grey}{0.50 / 0.08} \\
Gemma 4 31B       & R / P & 0.646 & 0.348 & 0.547 & 0.55 / 0.86 & 0.76 / 0.84 & \textcolor{grey}{0.50 / 0.11} & \textcolor{grey}{1.00 / 0.05} \\
Llama 3.3 70B     & R / P & 0.730 & 0.403 & 0.569 & 0.72 / 0.87 & 0.75 / 0.83 & \textcolor{grey}{0.50 / 0.08} & \textcolor{grey}{1.00 / 0.18} \\
\midrule
Majority vote & R / P & 0.646 & {0.345} & {0.533} & {0.55 / 0.86} & {0.76 / 0.84} & \textcolor{grey}{{0.50 / 0.08}} & \textcolor{grey}{{1.00 / 0.07}} \\
\midrule
\multicolumn{2}{@{}l}{\textit{vs.\ BARD}} & & & & (n=50) & (n=108) & (n=11) & (n=22) \\
\midrule
Claude Sonnet 4.6 & R / P & 0.855 & 0.642 & 0.810 & 0.93 / 0.93 & 0.81 / 0.99 & 1.00 / 0.44 & 0.84 / 0.73 \\
Gemini 3.6 Flash  & R / P & 0.874 & 0.659 & 0.810 & 0.94 / 0.92 & 0.83 / 0.99 & 1.00 / 0.42 & 0.86 / 0.86 \\
GPT-4o            & R / P & 0.817 & 0.561 & 0.727 & 0.96 / 0.86 & 0.81 / 0.99 & 0.55 / 0.30 & 0.68 / 0.58 \\
\midrule
GPT-OSS 20B       & R / P & 0.832 & 0.574 & 0.690 & 0.96 / 0.98 & 0.81 / 0.99 & 0.27 / 0.17 & 0.91 / 0.71 \\
Gemma 3 27B       & R / P & 0.785 & 0.455 & 0.718 & 0.96 / 0.67 & 0.88 / 0.99 & 0.45 / 0.42 & 0.09 / 0.18 \\
Gemma 4 31B       & R / P & 0.827 & 0.542 & 0.665 & 0.96 / 0.96 & 0.81 / 0.99 & 0.18 / 0.12 & 0.91 / 0.59 \\
Llama 3.3 70B     & R / P & 0.801 & 0.545 & 0.759 & 0.98 / 0.70 & 0.81 / 0.99 & 0.91 / 0.42 & 0.32 / 0.78 \\
\midrule
Majority vote & R / P & 0.864 & 0.637 & 0.789 & 0.96{ / }0.96 & 0.81 {/} 0.99 & 0.82{ / }0.38 & 0.91{ /} 0.74 \\
\bottomrule
\end{tabular}
\end{adjustbox}
\end{table}
\subsection{Extending to a Second BAO Axis: Detection Technology}
To test whether our auditing workflow generalizes beyond a single BAO axis, we repeated the analysis on the physical detection technology axis. BAO's detection hierarchy contains 12 classes, compared to 6 in assay format, at the depth-one level. Moreover, ChEMBL provides fewer cross-referenced labels, so we treat these results as a complementary probe. 

Table~\ref{tab:per_class_detection} reports the precision and recall per class. LLMs achieve high recall on most categories, including fluorescence, luminescence and radiometry. The performance on spectrophotometry is more moderate, with the mass spectrometry results and label-free methods having too few labels to draw definitive conclusions. Again, we see no big differences between open-source and proprietary models. 

To understand if the performance against silver labels is representative, and whether other questionable entries in PubChem might benefit from further review, we manually inspected 34 assays where the majority of LLMs disagreed with the label. BARD disagreed with the third-party annotation in the bulk download in all 24 cases where a BARD annotation was available. In 15 out of these 24 cases, our LLMs agreed with BARD, highlighting that the predictions that disagreed with the "silver" truth might be predominantly attributable to inaccuracies in the silver labels rather than genuine model error.

A recurring source of ambiguity was the occurrence of assay-supporting methods, such as "quantitative reverse transcription PCR" and "cell number determination," which do not map cleanly onto any BAO physical detection method subcategory. LLMs variably placed such assays under simple measurement, visual observation, or a physically-adjacent readout method (e.g., fluorescence, when the underlying chemistry, such as TaqMan, is fluorescent based), suggesting that forcing LLMs to output a single physical detection method is suboptimal for some assays.

 There was one AID (1259354)  with the notation ``AlphaLISA: fluorescence intensity'' in the assay dataset. We were unsure of whether AlphaLISA should be in the fluorescence methods, the luminescence method or the proximity-based method. All our LLMs tested agreed that it should be the luminescence method, which given the alphascreen method being in that category, could be reasonable.

To probe whether the remaining disagreements between BARD or ChEMBL and the LLMs could be explained, we asked the expert curator to review 11 records. In 6 cases the expert judged the assay genuinely undefinable and in every one of these, non-erroring models likewise self-reported low confidence, evidence that models can recognize when a physical detection method truly cannot be determined from the text, even as they still emit a best-guess label. In a further 3 cases (588778, 624286, 651578), the expert's own annotation explicitly named more than one acceptable detection method rather than a single ground truth: for 588778, the expert confirmed the assay genuinely uses both spectrophotometry (growth) and fluorescence (motility discrimination), while for 624286 and 651578 the expert accepted visual observation as one of several defensible labels for manual plaque counting. 

Only 2 of the 11 assays remain cases of true single-label disagreement: for AID 623918, the expert's chosen label (simple measurement) matched the model majority (4 of 7 models) against the label in BARD. For AID 623968, the expert's label (spectrophotometry) was tied 3–3 among non-error models against fluorescence, reflecting a genuine two-step-protocol ambiguity (RNA quantification by absorbance vs. RT-PCR detection via TaqMan fluorescence) rather than an outright model error. For the full analysis we refer to Appendix \ref{app:full_disagreement_table_detection}.

\begin{table}[t]
\centering
\caption{Per-class recall / precision on the BAO detection technology task.
Rare classes ($n<10$) in \textcolor{grey}{grey}.}
\label{tab:per_class_detection}
\begin{adjustbox}{width=\textwidth}
\begin{tabular}{@{}l cc cc c c c c@{}}
\toprule
\textbf{Model} & \textbf{Acc.} & \textbf{mF1} & \textbf{fluorescence} & \textbf{luminescence} & \textbf{spectrophotometry} & \textbf{radiometry} & \textcolor{grey}{\textbf{mass spec.}} & \textcolor{grey}{\textbf{label free}} \\
 & & & (n=141--159) & (n=67--80) & (n=54--55) & (n=10--12) & \textcolor{grey}{(n=1)} & \textcolor{grey}{(n=1)} \\
\midrule
Claude Sonnet 4.6 & 0.905 & 0.831 & 0.91 / 0.97 & 0.99 / 0.94 & 0.80 / 0.91 & 0.90 / 0.90 & \textcolor{grey}{1.00 / 0.20} & \textcolor{grey}{1.00 / 1.00} \\
Gemini 3.6 Flash  & 0.899 & 0.819 & 0.92 / 0.95 & 0.96 / 0.93 & 0.75 / 0.93 & 0.83 / 0.91 & \textcolor{grey}{1.00 / 0.20} & \textcolor{grey}{1.00 / 1.00} \\
GPT-4o            & 0.893 & 0.770 & 0.89 / 0.99 & 0.96 / 0.93 & 0.80 / 0.96 & 0.83 / 0.91 & \textcolor{grey}{1.00 / 0.20} & \textcolor{grey}{1.00 / 0.50} \\
\midrule
GPT-OSS 20B       & 0.883 & 0.817 & 0.89 / 0.97 & 0.96 / 0.93 & 0.75 / 0.93 & 0.83 / 0.91 & \textcolor{grey}{1.00 / 0.20} & \textcolor{grey}{1.00 / 1.00} \\
Gemma 3 27B       & 0.886 & 0.822 & 0.89 / 0.97 & 0.96 / 0.94 & 0.76 / 0.95 & 0.83 / 0.91 & \textcolor{grey}{1.00 / 0.20} & \textcolor{grey}{1.00 / 1.00} \\
Gemma 4 31B       & 0.892 & 0.719 & 0.92 / 0.97 & 0.95 / 0.94 & 0.75 / 0.93 & 0.83 / 0.91 & \textcolor{grey}{1.00 / 0.20} & \textcolor{grey}{1.00 / 0.25} \\
Llama 3.3 70B     & 0.883 & 0.817 & 0.89 / 0.96 & 0.96 / 0.93 & 0.76 / 0.91 & 0.83 / 0.91 & \textcolor{grey}{1.00 / 0.20} & \textcolor{grey}{1.00 / 1.00} \\
\midrule
{Majority vote} & {0.890} & {0.819} & {0.91 / 0.96} & {0.96 / 0.94} & {0.75 / 0.93} & {0.83 / 0.91} & \textcolor{grey}{{1.00 / 0.20}} & \textcolor{grey}{{1.00 / 1.00}} \\
\bottomrule
\end{tabular}
\end{adjustbox}
\end{table}

\subsection{Calibrated confidence and Prompt Ablations} 
\label{sec:trust}
Intrigued by the finding that all the LLMs labelled the assay which the expert labeled as ``cannot define'' with low confidence, we examined the relationship between LLM-reported confidence and correctness against the silver label (see Appendix~\ref{app:confidence_calibration}). For the 1097 ChEMBL datapoints, over all the seven LLMs tried, we saw that for the medium confidence labels given by Claude and GPT-OSS 20B, the disagreement rate at 33\% and 100\% was much higher than for the instances predicted high confidence and a disagreement rate of only up to 5\%. For the detection task we saw more spread confidence. On the high confidence labels there is a disagreement rate of up to only 10\%, with higher disagreement for the medium high labels and low confidence labels. It is also noteworthy that only Claude Sonnet 4.6 and GPT-OSS 20B model use the medium confidence, though other models give low confidence to some of their predictions as well. Combined with reasoning traces, this enables a practical triage: high-confidence predictions that agree with BARD can be accepted; high-confidence disagreements are surfaced with their supporting evidence for human review; and low-confidence predictions are routed to curators along with the model's reasoning. Nevertheless, further verification is necessary, especially since these are disagreement rates against silver labels.

Next to having a calibrated confidence, it is important that the LLMs predict based on the information given, not merely the order of the labels presented. To investigate the sensitivity of our results to prompt design, we evaluated
two perturbations of the baseline prompt on the ChEMBL assay format
subset. First the order swap where we reversed the alphabetical order of the initial experiment. Second, the format without definitions, which removed the BAO class definitions and only keeps the assay format labels. Results show that when swapping the order, labels change in less than 1.2\% of the cases for proprietary models and 2.6\% for the open-source models, indicating low ordering bias (see Appendix \ref{app:prompt_ablations}). 
Removing the definitions, however, changes 3.1\%, 10.8\%, and 3.8\% of predictions for
Claude Sonnet 4.6, Gemini 3.6 Flash, and GPT-4o respectively. Gemini in particular shows more changes away from the silver label, which implies that Gemini might rely more heavily on the in-prompt definitions. For the open-source models the change is 4.2\% for Llama 3.3 70B, up to 8.8\% for GPT-4o.  Interestingly, without the definitions there are also a number of flips without the definitions towards the silver label. This could mimic the human pattern of guessing the definition and consequently assigning it to a certain (silver) label, without grounding it in the exact definition asked for.

\section{Discussion and Conclusion}
\label{sec:discussion}
Our audit indicates that PubChem's bulk distribution offers incomplete assay metadata: roughly 36\% of records lack an assay format, and fewer than 0.1\% carry a structured BAO annotation. We also observe non-trivial disagreement across resources, 10\% among the 236 assays with both BARD and ChEMBL format labels available in the bulk download. These gaps likely reflect the optional nature of several metadata fields at deposition time, together with the limited vocabulary of PubChem's native Assay Format field relative to the richer BAO hierarchy.

 To address this gap, we evaluated seven LLMs, both proprietary and open-source, and found that they agree strongly with each other and recover well-supported assay format classes with recall above 0.96. Manual inspection of the cases where LLMs disagreed with silver labels showed that these errors are frequently traceable to inconsistencies between sources, or to mislabeling, rather than to the models themselves: in over a third of cases where the majority LLM prediction disagreed with ChEMBL, the LLMs in fact aligned with the PubChem/BARD label, and our senior annotator revised their own labels in several cases after seeing the model's cited evidence. Differences between proprietary and open-source models were small throughout, suggesting cheaper open-weight models are a viable option for annotation at scale, though at a cost of \$3--9 to audit 1000 records even proprietary models are orders of magnitude cheaper than manual re-curation(Appendix~\ref{app: resource usage}).

Taken together, results support a cautious claim that LLMs may improve upon individual existing annotations in cases of source disagreement. Based on this, we foresee a specific role for LLMs in the annotation pipeline, both as annotators and as auditors. LLMs flag inconsistent records, surface the textual evidence behind a disagreement, and route ambiguous cases to human curators, with per-class reliability estimates attached before any label is allowed to enter a downstream ML pipeline. 

These findings and the proposed framing come with limitations that follow from our results, silver labels are only partially reliable, so recall computed against them should be read as agreement with existing curation, not as ground-truth accuracy; where BARD, ChEMBL, and PubChem disagree, recall figures for the ``losing'' source will look artificially low even when the model is correct. Moreover, our evaluation stays at BAO depth one, and the live PUG JSON endpoint, the only consistently up-to-date source, currently rate-limits exhaustive crawls, which constrained how deep and how broad our dataset could go when sticking to PubChem. Our qualitative validation also has its own limitations: expert review was conducted with a single senior curator, so our findings reflect one expert's judgment rather than consensus across annotators. 

Finally, we identified one failure mode worth flagging explicitly: for AID 504631, several LLMs quoted the correct evidence (that \textit{Plasmodium falciparum} is a single-celled organism) yet still defaulted to a \textit{cell-based} label instead of \textit{organism-based}. This shows that even high-confidence predictions backed by correct evidence can fail at the final step of mapping that evidence onto the BAO class definitions, and argues for continued human spot-checking of whether the cited evidence actually supports the assigned label in these known tricky cases.

Beyond auditing existing annotations, the process itself surfaced gaps in BAO: the ontology currently lacks a definition for the ``biosensor'' detection technology, and the definitions of ``simple measurement'' and ``visual observation'' methods proved ambiguous enough to divide both LLMs and our expert curator. This suggests that LLM-based auditing can feed back into ontology maintenance, not only into data curation.

As next steps, we would like to experiment with deeper BAO granularity where support permits, expanding the dataset that we test on, utilize industrial datasets and LLM-mediated
debate to resolve residual silver-label conflicts, ingredients we believe necessary to bring legacy bioassay repositories to the readiness that foundation models for molecular property prediction now demand.

\paragraph{LLM disclosure.} LLMs were extensively used throughout this work. Primarily, as part of the experimental setup. LLMs were used for brainstorming, input for paper writing, and supported coding, mainly for plotting and analysis of the results. All written content was authored and edited by the authors.

\section{Acknowledgments}
This research was supported by the Wallenberg AI, Autonomous Systems and Software Program (WASP) funded by the Knut and Alice Wallenberg Foundation. Computational resources were provided on the Berzelius system funded by the Knut and Alice Wallenberg foundation and operated by NAISS in combination with resources provided by AstraZeneca.

\newpage

\printbibliography
\clearpage

\appendix

\section*{Appendix}

\section{Dataset Creation Details}
\label{app:dataset-creation}

To evaluate the performance of LLMs for predicting bioassay metadata
classes, we performed a bulk download from PubChem BioAssay of the
following resources: the JSON zip
files\footnote{\url{https://ftp.ncbi.nlm.nih.gov/PubChem/Bioassay/JSON/}},
\texttt{bioassays.tsv.gz}\footnote{\url{https://ftp.ncbi.nlm.nih.gov/PubChem/Bioassay/Extras/bioassays.tsv.gz}},
\texttt{Aid2CategorizedComment.gz}\footnote{\url{https://ftp.ncbi.nlm.nih.gov/PubChem/Bioassay/Extras/Aid2CategorizedComment.gz}},
and \texttt{Aid2Annotation.gz}\footnote{\url{https://ftp.ncbi.nlm.nih.gov/PubChem/Bioassay/Extras/Aid2Annotation.gz}}.

Seven of the JSON zip files, containing categorized bioassay names,
comments, and descriptions, were empty at the time of download:
\texttt{1258001\_1259000}, \texttt{1343001\_1344000},
\texttt{1344001\_1345000}, \texttt{1641001\_1642000},
\texttt{1642001\_1643000}, \texttt{1643001\_1644000}, and
\texttt{1644001\_1645000}. This is itself a small illustration of the
data-quality issues affecting PubChem's bulk distribution that motivate
this work.

In total, we successfully downloaded 1 994 310 BioAssay AIDs, of which
1 770 568 contained a name, a comment, and/or a description. To mitigate
this missing data, for the AIDs used in our analysis we additionally
queried PubChem's PUG REST
endpoint\footnote{\url{https://PubChem.ncbi.nlm.nih.gov/rest/pug_view/data/bioassay/\{aid\}/JSON}}
directly, which contained a description for all assays of interest.

\section{Dataset information from bulk PubChem}

Tables~\ref{tab:assay_fields},~\ref{tab:assay_formats},~\ref{tab:bao_assay_formats},~\ref{tab:bao_detection_technologies}, and~\ref{tab:bioassay_types} report the statistics of the data collected from the PubChem BioAssay database bulk download. Raw data can be downloaded from \url{https://ftp.ncbi.nlm.nih.gov/pubchem/Bioassay/}.

\begin{table}[h]

\centering
\caption{Assay coverage and uniqueness for description, comment, and name fields (Total AIDs = 1{ }994{ }310)}
\label{tab:assay_fields}
\resizebox{\textwidth}{!}{
\begin{tabular}{lrrrr}
\toprule
\textbf{Field} & \textbf{Non-empty} & \textbf{\% of Total} & \textbf{Unique Values} & \textbf{\% Unique (of non-empty)} \\
\midrule
AID & 1 994 310 & 100.0\% & 1 994 310 & 100.00\% \\
Comment & 1 769 245 & 88.7\% & 647 580 & 36.60\% \\
Description & 1 770 567 & 88.8\% & 173 040 & 9.77\% \\
Name & 1 770 568 & 88.8\% & 1 559 216 & 88.06\% \\
Deposit Date & 1 910 340 & 95.8\% & 1 979 & 0.10\% \\
Assay Format & 1 278 238 & 64.1\% & 3 & 0.00\% \\
Comment.BAO: assay format & 280 & 0.0\% & 8 & 2.50\% \\
Comment.BAO: detection technology & 348 & 0.0\% & 16 & 4.31\% \\
\bottomrule
\end{tabular}
}
\end{table}

\subsection{Dataset tables investigated labels statistics}

The tables~\ref{tab:assay_formats},~\ref{tab:bao_assay_formats},~\ref{tab:bao_detection_technologies}, and~\ref{tab:bioassay_types} investigate for each of the selected labels the coverage over which description, comment, or name are present.

\begin{table}[!htbp]
    \centering
    \caption{Distribution of assay formats available after bulk-downloading and processing PubChem BioAssay, broken down by commented vs. total counts and comment coverage. \emph{With Comment} counts assays that have a description, comment, or name; \emph{Total} counts all assays in the category; \emph{Coverage} is the annotation rate per category (With Comment / Total); \emph{Percentage} is each category's share of the commented subset.}
    \label{tab:assay_formats}
    \resizebox{0.65\textwidth}{!}{
    \begin{tabular}{lrrrr}
        \toprule
        \textbf{Assay Format} & \textbf{With Comment} & \textbf{Total} & \textbf{Coverage} & \textbf{Percentage} \\
        \midrule
        Organism-based & 609 321 & 663 381 & 91.85\% & 52.96\% \\
        Cell-based & 541 113 & 614 738 & 88.02\% & 47.03\% \\
        Biochemical & 101 & 119 & 84.87\% & 0.01\% \\
        \midrule
        \textbf{Total} & \textbf{1 150 535} & \textbf{1 278 238} & \textbf{90.01\%} & \textbf{100.00\%} \\
        \bottomrule
    \end{tabular}
    }
\end{table}

\begin{table}[htbp]
    \centering
    \caption{Distribution of BAO Assay Formats available after bulk-downloading and processing PubChem BioAssay, broken down by commented vs. total counts and comment coverage. \emph{With Comment} counts assays that have a description, comment, or name; \emph{Total} counts all assays in the category; \emph{Coverage} is the annotation rate per category (With Comment / Total); \emph{Percentage} is each category's share of the commented subset.}
    \label{tab:bao_assay_formats}
    \resizebox{\textwidth}{!}{
    \begin{tabular}{lrrrr}
        \toprule
        \textbf{BAO Assay Format} & \textbf{With Comment} & \textbf{Total} & \textbf{Coverage} & \textbf{Percentage} \\
        \midrule
        cell-based format & 112 & 119 & 94.12\% & 44.80\% \\
        biochemical format: protein format: single protein format & 91 & 103 & 88.35\% & 36.40\% \\
        biochemical format: protein format: protein complex format & 27 & 34 & 79.41\% & 10.80\% \\
        cell based format & 9 & 12 & 75.00\% & 3.60\% \\
        biochemical format: protein format: Single protein format & 5 & 6 & 83.33\% & 2.00\% \\
        organism-based format & 4 & 4 & 100.00\% & 1.60\% \\
        cell-free format: subcellular: cytosol format & 2 & 2 & 100.00\% & 0.80\% \\
        \midrule
        \textbf{Total} & \textbf{250} & \textbf{280} & \textbf{89.29\%} & \textbf{100.00\%} \\
        \bottomrule
    \end{tabular}
    }
\end{table}

\begin{table}[htbp]
    \centering
    \caption{Distribution of BAO Detection Technologies available after bulk-downloading and processing PubChem BioAssay, broken down by commented vs. total counts and comment coverage. \emph{With Comment} counts assays that have a description, comment, or name; \emph{Total} counts all assays in the category; \emph{Coverage} is the annotation rate per category (With Comment / Total); \emph{Percentage} is each category's share of the commented subset.}
    \label{tab:bao_detection_technologies}
    \resizebox{\textwidth}{!}{
    \begin{tabular}{lrrrr}
        \toprule
        \textbf{BAO Detection Technology} & \textbf{With Comment} & \textbf{Total} & \textbf{Coverage} & \textbf{Percentage} \\
        \midrule
        AlphaLISA: fluorescence intensity & 1 & 1 & 100.00\% & 0.31\% \\
        absorbance & 2 & 2 & 100.00\% & 0.63\% \\
        fluorescence: flow cytometry & 37 & 37 & 100.00\% & 11.64\% \\
        fluorescence: fluorescence intensity & 112 & 115 & 97.39\% & 35.22\% \\
        fluorescence: fluorescence polarization & 21 & 24 & 87.50\% & 6.60\% \\
        fluorescence: fret: htrf & 2 & 3 & 66.67\% & 0.63\% \\
        fluorescence: fret: tr-fret & 10 & 10 & 100.00\% & 3.14\% \\
        label free technology: isothermal titration calorimetry & 1 & 1 & 100.00\% & 0.31\% \\
        label free technology: mass spectrometry & 1 & 1 & 100.00\% & 0.31\% \\
        luminescence: bioluminescence & 1 & 1 & 100.00\% & 0.31\% \\
        luminescence: chemiluminescence & 79 & 86 & 91.86\% & 24.84\% \\
        radiometry: scintillation counting: filter assay & 2 & 6 & 33.33\% & 0.63\% \\
        radiometry: scintillation counting: scintillation proximity assay & 2 & 6 & 33.33\% & 0.63\% \\
        spectrophotometry: absorbance & 46 & 54 & 85.19\% & 14.47\% \\
        spectrophotometry: absorbance\_ $\mid$  $\mid$ \_fluorescence: fluorescence intensity & 1 & 1 & 100.00\% & 0.31\% \\
        \midrule
        \textbf{Total} & \textbf{318} & \textbf{348} & \textbf{91.38\%} & \textbf{100.00\%} \\
        \bottomrule
    \end{tabular}
    }
\end{table}

\begin{table}[htbp]
    \centering
    \caption{Distribution of BioAssay Types available after bulk-downloading and processing PubChem BioAssay, broken down by commented vs. total counts and comment coverage. \emph{With Comment} counts assays that have a description, comment, or name; \emph{Total} counts all assays in the category; \emph{Coverage} is the annotation rate per category (With Comment / Total); \emph{Percentage} is each category's share of the commented subset.}
    \label{tab:bioassay_types}
    \resizebox{\textwidth}{!}{
    \begin{tabular}{lrrrr}
        \toprule
        \textbf{BioAssay Type} & \textbf{With Comment} & \textbf{Total} & \textbf{Coverage} & \textbf{Percentage} \\
        \midrule
        Biochemical & 1 866 & 3 155 & 59.14\% & 0.90\% \\
        Biochemical $\mid$ Cell-based & 18 & 22 & 81.82\% & 0.01\% \\
        Biochemical $\mid$ Cell-based $\mid$ In vivo & 1 & 1 & 100.00\% & 0.00\% \\
        Biochemical $\mid$ Cell-based $\mid$ Toxicity & 2 & 3 & 66.67\% & 0.00\% \\
        Biochemical $\mid$ In vitro & 24 & 48 & 50.00\% & 0.01\% \\
        Biochemical $\mid$ In vivo & 68 & 68 & 100.00\% & 0.03\% \\
        Biochemical $\mid$ RNAi & 1 & 1 & 100.00\% & 0.00\% \\
        Biochemical $\mid$ Toxicity & 337 & 385 & 87.53\% & 0.16\% \\
        Cell-based & 3 578 & 4 184 & 85.52\% & 1.73\% \\
        Cell-based $\mid$ In vitro & 49 & 55 & 89.09\% & 0.02\% \\
        Cell-based $\mid$ In vivo & 2 & 2 & 100.00\% & 0.00\% \\
        Cell-based $\mid$ Organism-based & 27 & 39 & 69.23\% & 0.01\% \\
        Cell-based $\mid$ Organism-based $\mid$ Toxicity & 3 & 3 & 100.00\% & 0.00\% \\
        Cell-based $\mid$ RNAi & 52 & 52 & 100.00\% & 0.03\% \\
        Cell-based $\mid$ Toxicity & 289 & 347 & 83.29\% & 0.14\% \\
        In vitro & 25 224 & 26 604 & 94.81\% & 12.22\% \\
        In vitro $\mid$ In vivo & 30 & 53 & 56.60\% & 0.01\% \\
        In vitro $\mid$ Organism-based & 67 & 67 & 100.00\% & 0.03\% \\
        In vitro $\mid$ Toxicity & 1 644 & 1 658 & 99.16\% & 0.80\% \\
        In vivo & 22 663 & 23 965 & 94.57\% & 10.98\% \\
        In vivo $\mid$ Organism-based & 12 & 12 & 100.00\% & 0.01\% \\
        In vivo $\mid$ Organism-based $\mid$ Toxicity & 3 & 3 & 100.00\% & 0.00\% \\
        In vivo $\mid$ Toxicity & 255 & 266 & 95.86\% & 0.12\% \\
        Organism-based & 573 & 602 & 95.18\% & 0.28\% \\
        Organism-based $\mid$ Toxicity & 28 & 40 & 70.00\% & 0.01\% \\
        RNAi & 115 & 124 & 92.74\% & 0.06\% \\
        RNAi $\mid$ Toxicity & 7 & 8 & 87.50\% & 0.00\% \\
        Toxicity & 149 444 & 167 493 & 89.22\% & 72.41\% \\
        \midrule
        \textbf{Total} & \textbf{206 382} & \textbf{229 260} & \textbf{90.02\%} & \textbf{100.00\%} \\
        \bottomrule
    \end{tabular}
    }
\end{table}

\clearpage

\section{BAO format tree}
\label{app: BAO_tree}

\clearpage

\section{BAO detection tree}
\label{app: BAO_detection_tree}

\clearpage
\section{Prompts used }
\label{app: prompts_used}
Three prompt variants were evaluated for classifying bioassay records into
one of five BAO \emph{assay format} classes
(\emph{cell-based}, \emph{biochemical}, \emph{organism-based},
\emph{tissue-based}, \emph{cell-free}) using Claude via AWS Bedrock. We left out the small-molecule physicochemical format in the prompt as this was not part of our silver labels.
All variants share the same input schema
(\texttt{Title / Description / Comment / Protocol}) and the same JSON
output contract; they differ only in how the label and definitions are presented. For the physical detection method we used a prompt similar to the baseline.

\begin{itemize}[leftmargin=1.2em, itemsep=2pt]
  \item \textbf{Experiment 1 — Baseline:} full BAO definitions,
        of assay format, alphabetical class order.
  \item \textbf{Experiment 2 — Order swap:} full BAO definitions,
        \emph{reversed} class order (probes ordering bias).
  \item \textbf{Experiment 3 — Without definitions:} class names only,
        no definitions (probes reliance on label semantics vs.\ prior
        knowledge).
        \item \textbf{Experiment 4 — Physical Detection Method:} full BAO definitions of assay physical detection technology. 
\end{itemize}

\clearpage

\section*{Experiment 1 — baseline (with definitions)}
Presents each of the 5 classes with its official BAO definition and
subcategory list. Order: biochemical $\rightarrow$ cell-based
$\rightarrow$ cell-free $\rightarrow$ organism-based $\rightarrow$
tissue-based.

\begin{promptbox}[System prompt — \texttt{Baseline Assay Format Prompt}]{expBlue}
\small
You are an expert curator working with the BioAssay Ontology (BAO), a
structured vocabulary used by ChEMBL and PubChem to standardize how
bioassay experiments are described. The category definitions below are
taken directly from BAO's official class definitions.

\medskip
Given the \textbf{Description}, \textbf{Comment}, and \textbf{Protocol}
of a bioassay record (when available), classify its assay format into
exactly one of the following categories, based on the BioAssay
Ontology (BAO).

\medskip
\textbf{Categories (with definitions and typical phrasing you may see):}

\begin{enumerate}[leftmargin=1.4em, itemsep=3pt]
  \item \textbf{biochemical format} \newline
  An \emph{in vitro} format used to
        measure the activity of a biological macromolecule (purified
        protein or nucleic acid). Most often homogeneous, but can be
        heterogeneous when a solid phase (e.g.\ beads) immobilizes the
        macromolecule. Subcategories: \emph{protein format, nucleic acid
        format, protein complex format, single protein format}.
  \item \textbf{cell-based format}\newline
  Uses living cells of eukaryotic
        origin, derived from a multicellular organism; heterogeneous
        assay type. Typically permanent (immortal) cell lines, primary
        cells, or stem cells. Distinct from single-celled organism-based
        formats (e.g.\ bacteria, yeast). Subcategory: \emph{Transwell
        chamber format}.
  \item \textbf{cell-free format}\newline
  Originates from a cell but does not
        use intact (live) cells; excludes biochemical assays. Usually
        homogeneous; heterogeneous when components are bead-immobilized.
        Subcategories: \emph{subcellular, whole cell lysate, plasma,
        serum, blood, nucleosome, cell membrane, cytosol, microsome,
        mitochondrion, nuclear extract, lysate-based, liposome, S9
        fraction, tissue membrane, synaptosomal fraction, rabbit
        reticulocyte lysate, synaptosomal P2 fraction}.
  \item \textbf{organism-based format}\newline
  Uses a living organism, single
        or multi-celled; heterogeneous assay type.
  \item \textbf{tissue-based format}\newline Uses a tissue derived from a
        living organism; heterogeneous assay type.
\end{enumerate}

\medskip
\textbf{Rules:}
\begin{itemize}[leftmargin=1.4em, itemsep=1pt]
  \item Use the strongest evidence across \textbf{all} provided fields.
  \item If multiple categories seem plausible, pick the one most
        directly supported by \emph{explicit terminology} in the text,
        not inferred biology.
  \item Respond with \textbf{valid JSON only} — no markdown, no extra
        text.
\end{itemize}

\medskip
\textbf{Output format:}
\begin{verbatim}
{"bao_format": "<cell-based format|biochemical format|
                 organism-based format|tissue-based format|
                 cell-free format>",
 "confidence": "<high|medium|low>",
 "evidence": "<short quote or paraphrase from the text
              that supports your choice>"}
\end{verbatim}
\end{promptbox}

\section*{Experiment 2 - Order swap (with definitions, reversed order)}
Identical wording and definitions as Experiment 1, but the class list is
\emph{reversed}: tissue-based $\rightarrow$ organism-based $\rightarrow$
cell-free $\rightarrow$ cell-based $\rightarrow$ biochemical. Isolates
the effect of \emph{presentation order} on the model's choice.

\begin{promptbox}[System prompt — \texttt{order\_swap}]{expTeal}
\small
You are an expert curator working with the BioAssay Ontology (BAO)\ldots
\emph{(introductory paragraph identical to Experiment 1)}

\medskip
\textbf{Categories (with BAO definitions):}

\begin{enumerate}[leftmargin=1.4em, itemsep=3pt]
  \item \textbf{tissue-based format} — Uses a tissue derived from a
        living organism; heterogeneous assay type.
  \item \textbf{organism-based format} — Uses a living organism, single
        or multi-celled; heterogeneous assay type.
  \item \textbf{cell-free format} — Originates from a cell but does not
        use intact (live) cells; excludes biochemical assays. Full
        subcategory list identical to Experiment 1.
  \item \textbf{cell-based format} — Uses living cells of eukaryotic
        origin from a multicellular organism; heterogeneous. Immortal
        cell lines, primary cells, or stem cells. Distinct from
        single-celled organism-based formats.
  \item \textbf{biochemical format} — \emph{In vitro} format to measure
        activity of a purified biological macromolecule (protein or
        nucleic acid). Subcategories: \emph{protein, nucleic acid,
        protein complex, single protein}.
\end{enumerate}

\medskip
\textbf{Rules} and \textbf{Output format} are identical to Experiment 1.
\end{promptbox}

\section*{Experiment 3 — Without definitions (labels only)}
Strips out all BAO definitions and subcategory lists. The model sees
only the five class \emph{names} and must rely on its pretraining prior
about what each term means. Measures how much of the classifier's
performance depends on the definitions vs.\ label semantics.

\begin{promptbox}[System prompt — \texttt{results\_without\_definitions}]{expPlum}
\small
You are an expert curator working with the BioAssay Ontology (BAO)\ldots
\emph{(introductory paragraph identical to Experiment 1)}

\medskip
Given the \textbf{Description}, \textbf{Comment}, and \textbf{Protocol}
of a bioassay record (when available), classify its assay format into
exactly one of the following categories, based on the BioAssay
Ontology (BAO).

\medskip
\textbf{Categories:}
\begin{enumerate}[leftmargin=1.4em, itemsep=1pt]
  \item biochemical format
  \item cell-based format
  \item cell-free format
  \item organism-based format
  \item tissue-based format
\end{enumerate}

\medskip
\textbf{Rules:}
\begin{itemize}[leftmargin=1.4em, itemsep=1pt]
  \item Use the strongest evidence across \textbf{all} provided fields.
  \item If multiple categories seem plausible, pick the one most
        directly supported by \emph{explicit terminology} in the text,
        not inferred biology.
  \item Respond with \textbf{valid JSON only} — no markdown, no extra
        text.
\end{itemize}

\medskip
\textbf{Output format:} identical to Experiments 1 and 2.
\end{promptbox}

\newpage
\section*{Experiment 4 --- baseline (with definitions), detection method}
Presents each of the 12 BAO physical detection method classes with its
official definition and subcategory list. Order: biosensor
$\rightarrow$ fluorescence $\rightarrow$ imaging $\rightarrow$ isometric
tension recording $\rightarrow$ label free $\rightarrow$ luminescence
$\rightarrow$ mass spectrometry $\rightarrow$ proximity-based
$\rightarrow$ radiometry $\rightarrow$ simple measurement $\rightarrow$
spectrophotometry $\rightarrow$ visual observation.

\begin{promptbox}[System prompt --- \texttt{Baseline Physical Detection Method Prompt}]{expBlue}
\small
You are an expert curator working with the BioAssay Ontology (BAO), a
structured vocabulary used by ChEMBL and PubChem to standardize how
bioassay experiments are described. The category definitions below are
taken directly from BAO's official class definitions.

\medskip
Given the \textbf{Description}, \textbf{Comment}, and \textbf{Protocol}
of a bioassay record (when available), classify its \emph{physical
detection method} into exactly one of the following categories, based
on the BioAssay Ontology (BAO).

\medskip
\textbf{Categories (with definitions):}

\begin{enumerate}[leftmargin=1.4em, itemsep=3pt]
  \item \textbf{biosensor method}\newline
        This category has no subcategories.

  \item \textbf{fluorescence method}\newline
        Detection techniques that use the principles of fluorescence,
        whereby incident light excites a fluorophore which then emits
        light at lower energy (higher wavelength). The emitted light
        is typically from the visible portion of the UV--Visible
        spectrum. Subcategories: \emph{fluorescence resonance energy
        transfer, fluorescence polarization, flow cytometry, thermal
        shift, fluorescence intensity, homogeneous time resolved
        fluorescence, time resolved fluorescence resonance energy
        transfer}.

  \item \textbf{imaging method}\newline
        The field of using microscopes to see objects that cannot be
        seen with the unaided eye. Subcategories: \emph{tomography,
        microscopy, electron microscopy, optical microscopy, scanning
        probe microscopy, scanning electron microscopy, transmission
        electron microscopy, fluorescence microscopy, confocal
        microscopy, brightfield microscopy, phase contrast microscopy,
        dark field microscopy, differential interference contrast
        microscopy, scanning near-field optical microscopy, vibrational
        spectroscopic imaging, atomic force microscopy, scanning
        tunneling microscopy, total internal reflection fluorescence
        microscopy, Raman imaging, infrared imaging, near-infrared
        imaging, mid-infrared imaging}.

  \item \textbf{isometric tension recording method}\newline
        Isometric tension recording is a technique used to measure
        tension. This category has no subcategories.

  \item \textbf{label free method}\newline
        Allows detection of binding interactions and cell-based
        reactions without conventional labels (e.g.\ fluorescent
        probes). Advantages: (a) functional activity without modifying
        binding partners with labels, (b) binding interactions
        independent of functional activity, and (c) cell-based GPCR
        assays without engineering cell lines to over-express
        receptors. Subcategories: \emph{optical based, nuclear
        magnetic resonance, circular dichroism, electrical sensor,
        x-ray crystallography, isothermal titration calorimetry,
        quartz crystal microbalance, surface plasmon resonance, bio
        layer interferometry, fiber optic waveguide, optical waveguide
        grating, resonant waveguide grating, microelectrode
        measurement, carbon nanotube based sensor, current clamp,
        patch clamp, voltage clamp}.

  \item \textbf{luminescence method}\newline
        Emission of light from a substance from an electronically
        excited state reached by a physical, mechanical, or chemical
        mechanism. Subcategories: \emph{bioluminescence,
        chemiluminescence, AlphaScreen, bioluminescence resonance
        energy transfer, chemiluminescence resonance energy transfer}.

  \item \textbf{mass spectrometry}\newline
        Used to detect compounds bound to protein targets; unbound
        compounds are separated by affinity-based, size-exclusion, or
        ultrafiltration methods, then bound compounds are eluted from
        the protein and detected by MS. Throughput varies with the
        separation method (typically 200--2700 compounds per run).
        Also used to determine the chemical structure of molecules:
        samples are vaporized, ionized, and the mass-to-charge ratio
        of the particles is analyzed. This category has no
        subcategories.

  \item \textbf{proximity-based method}\newline
        A biophysical approach used to monitor the spatial closeness
        of proteins or molecules within live cells by exploiting
        measurable signals generated through interactions (e.g.\
        energy transfer) between tagged molecular components. For
        example, Bioluminescence Resonance Energy Transfer (BRET)
        uses energy transfer from a donor enzyme (e.g.\ luciferase)
        to an acceptor fluorophore, giving a quantifiable proximity
        signal. Widely applied to protein--protein interactions,
        protein--ligand interactions, receptor proximity, and
        intracellular signaling. Advances such as NanoBRET and
        NanoBiT have expanded scope further. This category has no
        subcategories.

  \item \textbf{radiometry method}\newline
        Detection technology using radioactive tracers (e.g.\ filter
        assay, SPA). Subcategories: \emph{scintillation counting,
        radiometric imaging method, Cherenkov counting method, gamma
        counting method, autoradiography, phosphorimaging}.

  \item \textbf{simple measurement method}\newline
        Any method entailing basic measurements requiring no special
        instrumentation (e.g.\ measuring tumor size with calipers, or
        a subject's weight with a scale). This category has no
        subcategories.

  \item \textbf{spectrophotometry method}\newline
        Uses a spectrophotometer to measure the amount of light a
        sample absorbs: a beam of light is passed through the sample
        and the intensity of light reaching a detector is measured.
        Subcategories: \emph{absorbance, transmittance, atomic
        absorption spectrophotometry}.

  \item \textbf{visual observation method}\newline
        Any method entailing primarily routine examination, requiring
        no special instrumentation. This category has no subcategories.
\end{enumerate}

\medskip
\textbf{Rules:}
\begin{itemize}[leftmargin=1.4em, itemsep=1pt]
  \item Use the strongest evidence across \textbf{all} provided fields.
  \item If multiple categories seem plausible, pick the one most
        directly supported by \emph{explicit terminology} in the text,
        not inferred biology.
  \item Respond with \textbf{valid JSON only} --- no markdown, no extra
        text.
\end{itemize}

\medskip
\textbf{Output format:}
\begin{verbatim}
{"bao_format": "<biosensor method|fluorescence method|
                    imaging method|isometric tension recording method|
                    label free method|luminescence method|
                    mass spectrometry|proximity-based method|
                    radiometry method|simple measurement method|
                    spectrophotometry method|visual observation method>",
 "confidence": "<high|medium|low>",
 "evidence": "<short quote or paraphrase from the text
              that supports your choice>"}
\end{verbatim}

\end{promptbox}

\section*{Shared user-turn template}

All experiments use the same per-record user message; missing
fields are replaced with the literal marker \texttt{(not provided)} so
the model is never shown \texttt{nan} or a silently blank line.

\begin{promptbox}[User prompt template (shared across all experiments)]{black!70}
\small
\begin{verbatim}
Title:       $record_title
Description: $description
Comment:     $comment
Protocol:    $protocol
\end{verbatim}
\end{promptbox}

\clearpage
\subsection{Case studies of annotation inconsistencies in PubChem}\label{app:annotation-inconsistencies}

To illustrate the annotation inconsistencies summarized in Section~\ref{sec: Missingness_inconsistencies}, in this section we discuss two representative case studies provide further statistics on the BARD-ChEMBL cross-source disagreements.

\paragraph{Case 1: Conflicting assay format labels (AID 1885).} AID 1885 is reported as \textit{Cell-based} in the PubChem web interface, but as \textit{Cell-based|Organism-based} in the bulk download. Although it is difficult to determine exactly how BioAssay Types are annotated, the protocol and description could plausibly support both categories: the assay runs in mouse fibroblasts (suggesting a cell-based format), while the target is a living microorganism (which could suggest an organism-based format). According to the BAO, however, the \textit{organism-based format} label refers only to intact living organisms\footnote{\BAOorganismURL}, so it should not apply here since no intact organism was used. The PubChem web interface correctly displays only the cell-based annotation. The same ambiguity likely affects the other 42 entries carrying \textit{Cell-based|Organism-based} annotations in PubChem, as well as many further combinations such as \textit{In Vitro|Organism-based}.

\paragraph{Case 2: Disagreement between third-party sources (AID 588585).} Next to annotations from external sources, PubChem also displays assay annotations obtained from third-party sources, and these do not always agree with each other or with the BARD annotations. Furthermore, the BARD-linked bulk download\footnote{See \url{https://pubchem.ncbi.nlm.nih.gov/source/BioAssay\%20Research\%20Database\%20(BARD)}.} does not cover all BARD-annotated assays shown on PubChem. AID 588585 presents an example for this. It is labelled ``biochemical'' under the categorised-comment BAO field but ``subcellular'' in the BARD-derived BioAssay Annotations tab, despite not appearing in the BARD bulk download at all. Across the JSON files corresponding to AIDs in the BARD bulk download, we found only one third-party annotated BAO label, on which BARD and the third party did agree.

\paragraph{Agreements on BARD--ChEMBL labels} In the bulk download of BARD we found 236 assays with both BARD and ChEMBL assay format labels on PubChem. Out of these, on 24 (10.2\%) assays, the two sources disagreed on the label. The disagreements comprised:

12 cases of \textit{Cell-based} vs. \textit{Organism-based};
6 cases of \textit{Biochemical} vs. \textit{Cell-based};
3 cases of \textit{Tissue-based} vs. \textit{Cell-based};
remaining cases including \textit{small-molecule physicochemical} (BARD) vs. \textit{Biochemical} (ChEMBL).
Together, these observations show that even highly manual, expert-driven curation does not guarantee consistency, either within a single record or across sources.
\clearpage
\section{Model Execution Parameters}
\label{app: model_params}

This appendix covers the exact runtime configuration used for each of
the seven models evaluated in this work. Any response whose predicted label fell outside the allowed BAO
class set was flagged as invalid.

\subsection{Structured output}

To make outputs comparable across providers, JSON output was enforced
at the API level wherever supported:

\begin{itemize}[leftmargin=1.4em, itemsep=1pt]
  \item \textbf{GPT-4o} (Azure OpenAI) and \textbf{Gemini 3.6 Flash}
        (Vertex AI, OpenAI-compatible passthrough): \\
        \texttt{response\_format=\{"type": "json\_object"\}}.
  \item \textbf{Claude Sonnet 4.6} (AWS Bedrock \texttt{Converse} API):
        no native JSON-mode flag is exposed by the Converse API, so JSON
        adherence was obtained purely via prompt instructions
        and parsed.
  \item \textbf{Open-weights models} (\texttt{Ollama v0.17.4}): a
        JSON-Schema was passed via the \texttt{format} parameter in such a way that the model literally cannot emit a
        label outside the allowed vocabulary.
\end{itemize}

\subsection{Proprietary models (API-served)}

\begin{table}[h]
\centering
\small
\begin{tabular}{lp{2.8cm}p{2.8cm}p{2.8cm}}
\toprule
\textbf{Parameter} & \textbf{GPT-4o} & \textbf{Gemini 3.6 Flash} & \textbf{Claude Sonnet 4.6}\\
\midrule
Provider / route      & Azure OpenAI                     & Vertex AI (via AI Gateway)         & AWS Bedrock (via AI Gateway) \\
Endpoint / API        & \texttt{chat.completions}        & OpenAI-compatible passthrough      & \texttt{bedrock-runtime.converse} \\
Model / deployment ID & \texttt{gpt-4o}                  & \texttt{google/ gemini-3.6-flash}   & \texttt{us.anthropic.\allowbreak claude-sonnet-4-6} \\
Temperature           & 0.0                              & 0.0                                & 0.0 \\
Max output tokens     & provider default                 & provider default                   & provider default \\
JSON-mode enforced?   & yes (\texttt{json\_object})      & yes (\texttt{json\_object})        & no (prompt-only) \\
\bottomrule
\end{tabular}
\caption{Runtime configuration for the three proprietary
API-served models. Temperature was fixed at \textbf{0.0} across all
models to obtain deterministic (greedy) decoding for reproducibility.
Maximum output length was left at the provider default: the response is
a compact JSON object with three short fields, well below any provider
cap.}
\label{tab:proprietary_params}
\end{table}

\subsection{Open-weights models (\texttt{Ollama v0.17.4})}

The open-weights models were served locally on the national
cluster. Each SLURM job started a dedicated
\texttt{ollama serve} process, ran the classifier via
\texttt{ollama.chat()}, and logged the GPU power draw once per second via \texttt{nvidia-smi} for downstream energy/CO$_{2}$ accounting
(see Appendix~\ref{app: resource usage}). All calls used a fixed
\texttt{seed=42} to make decoding reproducible.

\begin{table}[h]
\centering
\small
\begin{tabular}{lcccc}
\toprule
\textbf{Parameter} & \textbf{Gemma 4 31B} & \textbf{GPT-OSS 20B} & \textbf{Gemma 3 27B} & \textbf{Llama 3.3 70B}\\
\midrule
Ollama model tag        & \texttt{gemma4:31b} & \texttt{gpt-oss:20b} & \texttt{gemma3:27b}   & \texttt{llama3.3:70b} \\
Context window (\texttt{num\_ctx})     & 9000     & 9000     & 9000     & 9000 \\
Max output (\texttt{num\_predict})     & 8192     & 8192     & 8192     & 8192 \\
Temperature             & 0.0 & 0.0     & 0.0 & 0.0\\
Seed                    & 42       & 42       & 42       & 42 \\
Structured output       & JSON-Schema & JSON-Schema & JSON-Schema & JSON-Schema \\
\texttt{OLLAMA\_NUM\_PARALLEL} & 4  & 4        & 4        & 4 \\
\texttt{OLLAMA\_CONTEXT\_LENGTH} & 8192 & 8192  & 8192     & 8192 \\
\bottomrule
\end{tabular}
\caption{Runtime configuration for the four open-weights models served
via \texttt{Ollama v0.17.4}. All open-weights runs used
JSON-Schema-constrained decoding (Ollama's \texttt{format=} parameter),
which enforces the enum of valid BAO labels at generation time.}
\label{tab:open_weights_params}
\end{table}

\clearpage
\section{Inter-model agreement (Cohen's \texorpdfstring{$\kappa$}{kappa})}\label{app:cohen}

For each of the three classification tasks, the BAO assay format on the
ChEMBL evaluation set, BAO assay format on the PubChem-BAO subset, and
BAO detection technology, we computed the pairwise Cohen's $\kappa$
between all evaluated LLMs. Cohen's $\kappa$ measures agreement between
two annotators beyond what would be expected by chance, with $\kappa = 1$
indicating perfect agreement and $\kappa = 0$ indicating chance-level
agreement. Values above~0.8 are typically considered ``high''
agreement and values between 0.6 and 0.8 ``substantial'' agreement.
It is defined as: 
$\kappa = \frac{\Pr(a) - \Pr(e)}{1 - \Pr(e)}$, where $\Pr(a)$ is the observed proportion of agreement between the two annotators and $\Pr(e)$ is the proportion of agreement expected under chance~\cite{cohen1960coefficient}. We utilized \texttt{cohen\_kappa\_score} from \texttt{scikit-learn's metrics} to calculate the Cohen's $\kappa$ value. 

\begin{figure}[htbp]
    \centering
    \begin{minipage}{0.48\textwidth}
        \centering
        \includegraphics[width=\linewidth,trim=0 0 0 2cm, clip]{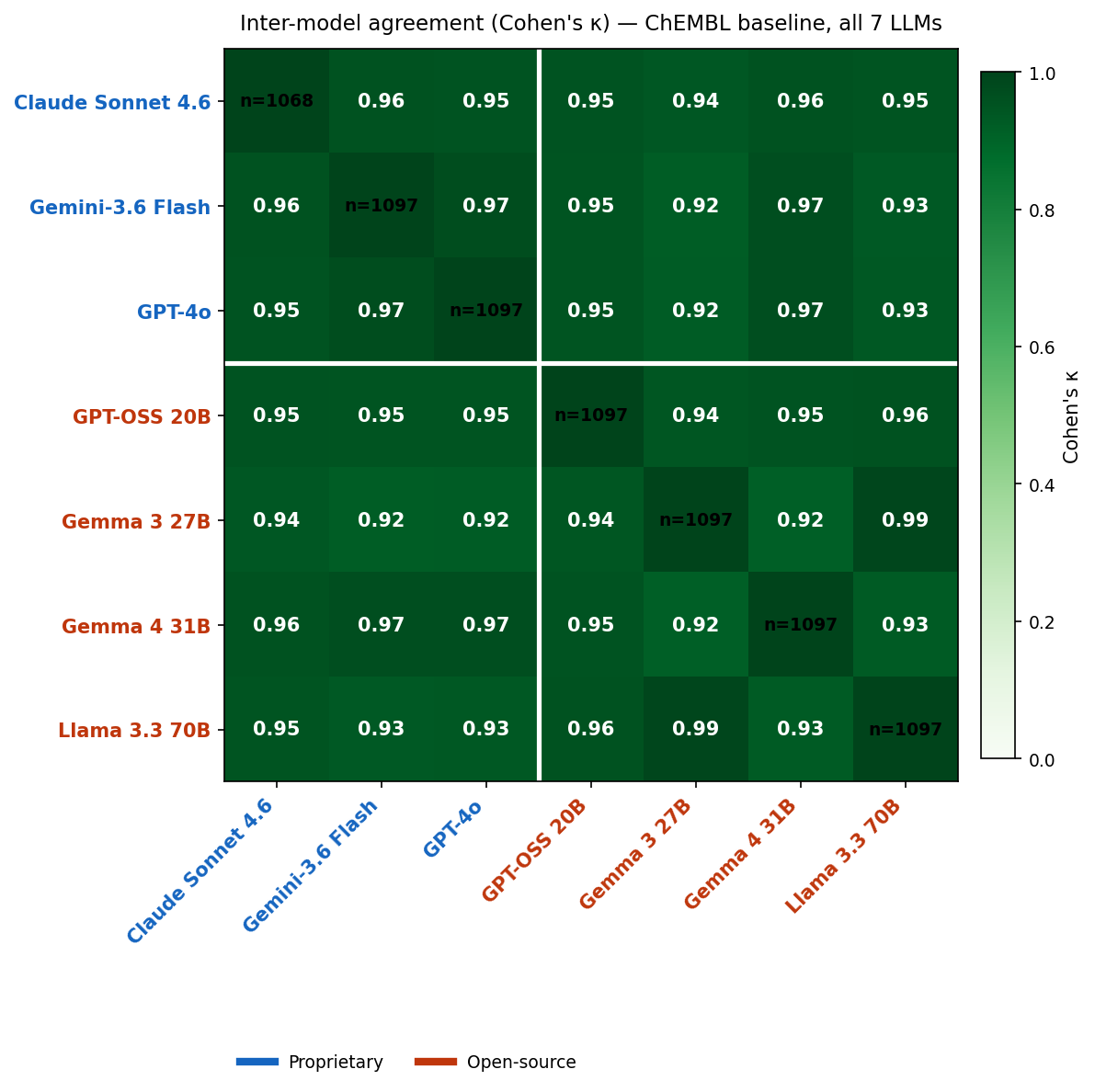}
        \caption{Pairwise Cohen's $\kappa$ between the seven evaluated
        LLMs on the ChEMBL assay format benchmark ($n=1097$). All
        model pairs exceed $\kappa > 0.9$, indicating near-perfect
        agreement across proprietary and open-weight models.}
        \label{fig:cohen_agreement}
    \end{minipage}
    \hfill
    \begin{minipage}{0.48\textwidth}
        \centering
        \includegraphics[width=\linewidth,trim=0 0 0 2cm, clip]{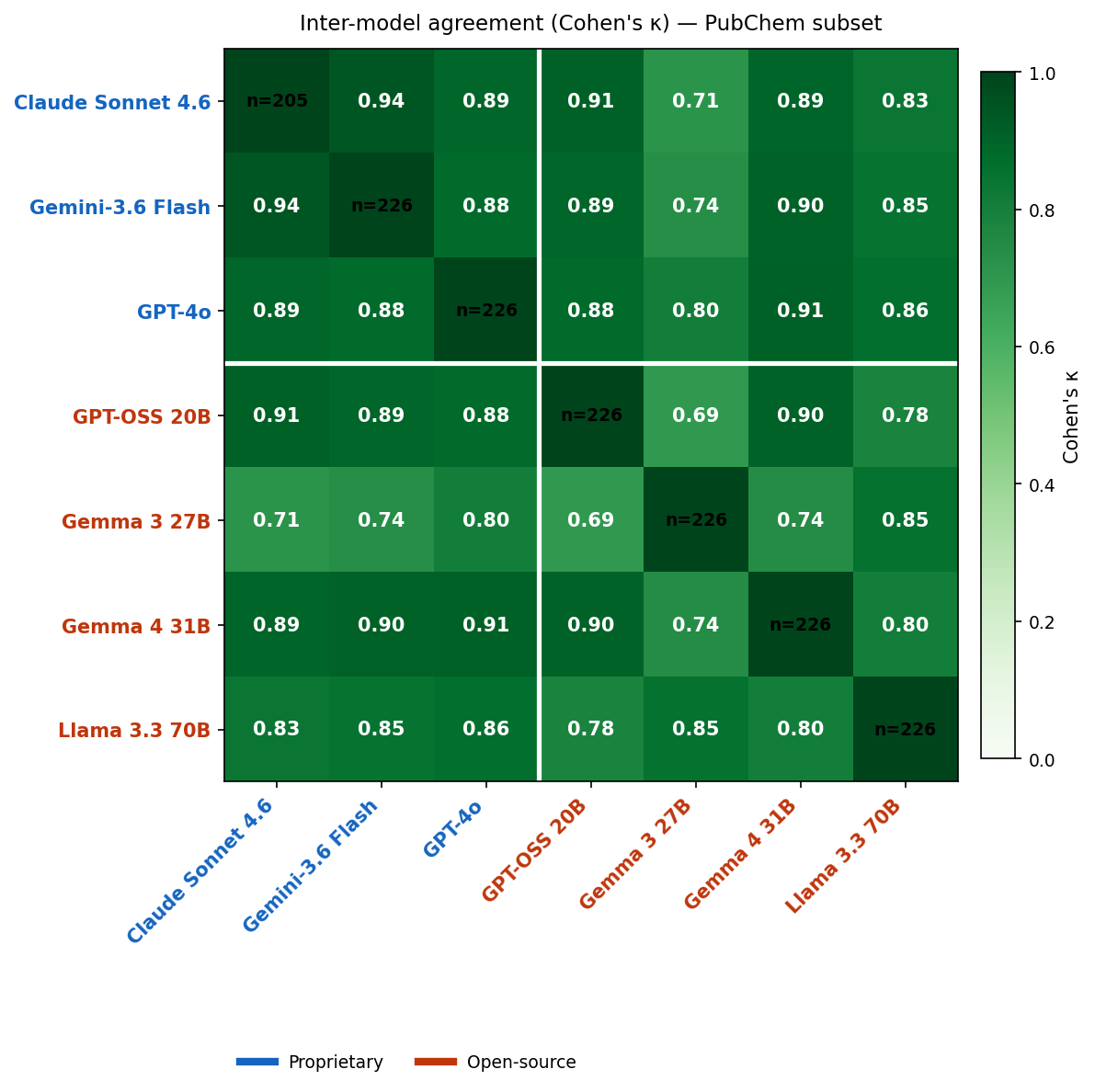}
        \caption{Pairwise Cohen's $\kappa$ between the seven LLMs on
        the smaller PubChem-BAO subset ($n=226$). Agreement is
        substantial ($\kappa > 0.8$ for most pairs) but lower than on
        the ChEMBL benchmark; disagreements concentrate on the
        organism-based and cell-free classes.}
        \label{fig:cohen_kappa_PubChem}
    \end{minipage}
\end{figure}

\begin{figure}[htbp]
    \centering
    \includegraphics[width=0.5\linewidth,trim=0 0.2cm 0 2cm, clip]{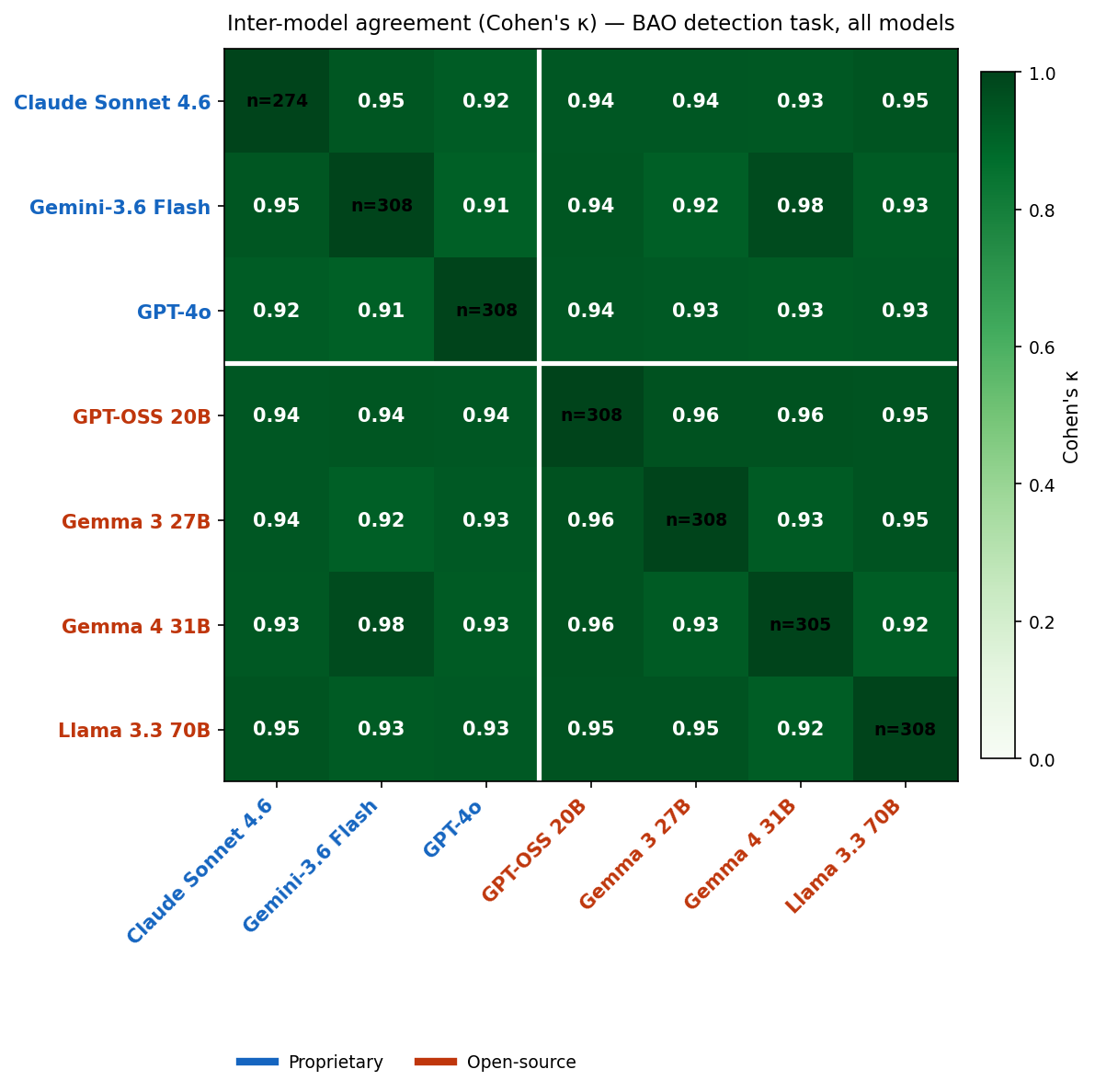}
    \caption{Pairwise Cohen's $\kappa$ between the three proprietary
    LLMs (Claude Sonnet~4.6, Gemini 3.6 Flash, GPT-4o) and four open source LLMs (GPT-OSS 20B, Gemma 3 27B, Gemma 4 41B, Llama 70B) on the BAO
    detection technology task. Agreement patterns mirror the
    assay format results: near-perfect agreement.}
    \label{fig:Cohen_kappa_detection}
\end{figure}

\clearpage

\section{Full manual-inspection table for ChEMBL assay format disagreements}
\label{app:full_disagreement_table}

For reproducibility, we provide the full row-level annotation of the 43
assays discussed in Section~\ref{sec:experiments}, including PubChem
AIDs, both silver-standard labels, the top-predicted LLM label, whether
a BARD annotation was available, and the vote count of the seven
evaluated models. Table~\ref{tab:Wrong_ChEMBL_Label}.

\begingroup
\small
\setlength{\tabcolsep}{4pt}
\renewcommand{\arraystretch}{1.15}

\begin{longtable}{@{}r L{1.5cm} L{2.1cm} L{1.6cm} L{4.3cm} r r@{}}
\caption{Manual inspection of the 43 assays where the majority of LLMs
disagreed with the ChEMBL BAO assay format label. All labels are BAO
assay formats; the ``\ format'' suffix is omitted for space. $n_\text{dis.}$ counts models disagreeing
with ChEMBL, $n_\text{al.}$ counts models agreeing with ChEMBL; seven models
were attempted per row unless noted otherwise. Rows shaded in gray were sent for,
and received, senior industrial curator review (Table~\ref{tab:Expert_Review}).}
\label{tab:Wrong_ChEMBL_Label} \\

\toprule
\textbf{AID} & \textbf{ChEMBL} & \textbf{BARD (if diff.)} &
\textbf{Top pred.} & \textbf{Comment} &
\textbf{$n_\text{dis.}$} & \textbf{$n_\text{al.}$} \\
\midrule
\endfirsthead

\multicolumn{7}{c}{\tablename\ \thetable\ -- \textit{continued from previous page}} \\
\toprule
\textbf{AID} & \textbf{ChEMBL} & \textbf{BARD (if diff.)} &
\textbf{Top pred.} & \textbf{Comment} &
\textbf{$n_\text{dis.}$} & \textbf{$n_\text{al.}$} \\
\midrule
\endhead

\midrule
\multicolumn{7}{r}{\textit{continued on next page}} \\
\endfoot

\bottomrule
\endlastfoot
1984  & biochemical & cell-based & cell-based & LLMs support BARD label & 7 & 0 \\
2121  & biochemical & cell-based & cell-based & LLMs support BARD label & 7 & 0 \\
2217  & cell-based & biochemical & biochemical & LLMs support BARD label & 7 & 0 \\
2613  & tissue-based & whole-cell lysate (subclass of cell-free) & cell-free & LLMs support BARD label & 7 & 0 \\
588382  & organism-based & cell-based & cell-based & LLMs support BARD label & 7 & 0 \\
588766  & cell-free & single-protein (part of biochemical) & biochemical & LLMs support BARD label & 7 & 0 \\
588780  & tissue-based & cell-based & cell-based & LLMs support BARD label & 7 & 0 \\
602210  & tissue-based & cell-based & cell-based & LLMs support BARD label & 7 & 0 \\
602211  & tissue-based & cell-based & cell-based & LLMs support BARD label & 7 & 0 \\
2592  & tissue-based & whole-cell lysate (part of cell-free) & cell-free & LLMs support BARD label & 7 & 0 \\
\rowcolor{gray!15}
1470  & biochemical & cell-free & cell-free & LLMs support BARD label & 6 & 1 \\
2588  & tissue-based & whole-cell lysate (part of cell-free) & cell-free & LLMs support BARD label & 6 & 1 \\
\rowcolor{gray!15}
488745  & organism-based & cell-based & cell-based & LLMs support BARD label & 5 & 2 \\
\rowcolor{gray!15}
488752  & organism-based & cell-based & cell-based & LLMs support BARD label & 5 & 2 \\
504546  & cell-based & biochemical & biochemical & LLMs support BARD label & 5 & 2 \\
504832  & organism-based & cell-based & cell-based & LLMs support BARD label & 5 & 2 \\
\rowcolor{gray!15}
2467  & cell-based & organism-based & organism-based & LLMs support BARD label & 4 & 3 \\

\rowcolor{gray!15}
492958  & cell-based & cell-based & organism-based & ChEMBL and BARD agree; LLMs disagree & 6 & 1 \\
\rowcolor{gray!15}
588769  & cell-free & cell-free & biochemical & ChEMBL and BARD agree; LLMs disagree & 6 & 1 \\
\rowcolor{gray!15}
1913  & biochemical & protein (subclass of biochemical) & cell-free & ChEMBL and BARD agree; LLMs disagree & 5 & 2 \\
\rowcolor{gray!15}
2693  & cell-based & cell-based & organism-based & ChEMBL and BARD agree; LLMs disagree & 5 & 2 \\
\rowcolor{gray!15}
488788  & biochemical & single-protein (part of biochemical) & cell-free & ChEMBL and BARD agree; LLMs disagree & 5 & 2 \\
\rowcolor{gray!15}
492957  & cell-based & cell-based & organism-based & ChEMBL and BARD agree; LLMs disagree & 5 & 2 \\
\rowcolor{gray!15}
2096  & cell-based & cell-based & organism-based & ChEMBL and BARD agree; LLMs disagree & 4 & 3 \\
\rowcolor{gray!15}
2695  & cell-based & cell-based & organism-based & ChEMBL and BARD agree; LLMs disagree & 4 & 3 \\
\rowcolor{gray!15}
492959  & cell-based & cell-based & organism-based & ChEMBL and BARD agree; LLMs disagree & 4 & 3 \\
\rowcolor{gray!15}
651893  & cell-based & cell-based & organism-based & ChEMBL and BARD agree; LLMs disagree & 4 & 3 \\
\rowcolor{gray!15}
504494  & biochemical & cell-based & cell-free & LLMs disagree with both ChEMBL and BARD & 5 & 2 \\
\rowcolor{gray!15}
463208  & biochemical & cell-based & cell-free (6 att.) & LLMs disagree with both ChEMBL and BARD & 4 & 2 \\

651720  & cell-based & -- & biochemical & -- & 7 & 0 \\
1159512  & organism-based & -- & cell-based & -- & 7 & 0 \\
449765  & organism-based & -- & cell-based & -- & 7 & 0 \\
651733  & cell-based & -- & biochemical & -- & 7 & 0 \\
652150  & cell-based & -- & organism-based & -- & 6 & 1 \\
1295  & cell-based & -- & cell-free & -- & 5 & 2 \\
504507  & biochemical & -- & cell-free & -- & 5 & 2 \\
743295  & cell-based & -- & biochemical & -- & 5 & 2 \\
1159566  & cell-based & -- & organism-based & -- & 5 & 2 \\
1159571  & cell-based & -- & organism-based & -- & 5 & 2 \\
\rowcolor{gray!15}
504631  & organism-based & -- & cell-based & -- & 4 & 3 \\
652151  & cell-based & -- & organism-based & -- & 4 & 3 \\
1159567  & cell-based & -- & organism-based & -- & 4 & 3 \\
1159570  & cell-based & -- & organism-based & -- & 4 & 3 \\

\end{longtable}
\endgroup

\begin{landscape}
\setlength{\tabcolsep}{4pt}
\renewcommand{\arraystretch}{1.2}

\begin{longtable}{@{}r L{0.08\linewidth} L{0.08\linewidth} L{0.21\linewidth} L{0.09\linewidth} L{0.20\linewidth} L{0.21\linewidth}@{}}
\caption{Assays sent for expert review, with ChEMBL label, top LLM prediction, and expert annotation.}
\label{tab:Expert_Review} \\

\toprule
\textbf{AID} & \textbf{ChEMBL} & \textbf{Top pred.} & \textbf{Short title} & \textbf{Expert format} & \textbf{Expert reason} & \textbf{Comment}\\
\midrule
\endfirsthead

\multicolumn{7}{c}{\tablename\ \thetable\ -- \textit{continued from previous page}} \\
\toprule
\textbf{AID} & \textbf{ChEMBL} & \textbf{Top pred.} & \textbf{Short title} & \textbf{Expert format} & \textbf{Expert reason} & \textbf{Comment}\\
\midrule
\endhead

\midrule
\multicolumn{7}{r}{\textit{continued on next page}} \\
\endfoot

\bottomrule
\endlastfoot

1470 & biochemical & cell-based & Discovery of novel allosteric modulators of the M1 muscarinic receptor: Agonist NMS binding at M1 & cell-free, but could also be biochemical & In protocol: ``Membranes were prepared from M1-expressing CHO cells''; format appears to be cell membranes but could be classified as biochemical as well & Expert finds multiple labels plausible\\

492958 & cell-based & organism-based & Counterscreen for AddAB inhibitors: absorbance-based bacterial cell-based high throughput dose response assay for inhibitors of bacterial viability & organism-based & see short title; you might even classify it as ``organism based'' since these are bacteria cells: later adapted when we discussed definition & Expert agrees with LLM majority: \textcolor{softgreen}{ChEMBL/BARD label incorrect} (expert initially leaned towards ChEMBL, but adapted after discussion) \\

588769 & cell-free & biochemical & Late stage assay provider results from the probe development effort to identify inhibitors of plasma platelet activating factor acetylhydrolase (pPAFAH): fluorescence-based dose response biochemical gel-based competitive Activity-Based Protein Profiling (ABPP) assay for HTS compounds & biochemical & fluorescence-based dose response biochemical gel-based competitive Activity-Based Protein Profiling (ABPP) assay for HTS compounds & Expert agrees with LLM majority: \textcolor{softgreen}{ChEMBL/BARD label incorrect}\\

1913 & biochemical & cell-free & Luminescence-based dose response biochemical high throughput screening assay for inhibitors of the Heat Shock Protein 90 (HSP90) & biochemical & see short title (it is using a ``reticulocyte lysate'' so classifying it as ``cell-free'' makes some sense as well) & Expert agrees with ChEMBL: \textcolor{softorange}{LLM majority plausible} \\

2693 & cell-based & organism-based & Fluorescence Cell-Based Dose Screen to Determine Inhibitors of S. cerevisiae Viability & organism-based & see short title (but could be defined as organism based since it is an organism, S. cerevisiae), but later adapted when we discussed definition & Expert agrees with LLM majority: \textcolor{softgreen}{ChEMBL/BARD label incorrect} (expert initially leaned towards ChEMBL, but adapted after discussion) \\

488745 & organism-based & cell-based & Quantitative high throughput screen for delayed death inhibitors of the malarial parasite plastid, 96 hour incubation & cell-based & in protocol: ``\ldots\ Four microliters of infected erythrocytes \ldots'' & Expert agrees with LLM majority: \textcolor{softgreen}{ChEMBL label incorrect (LLMs agree with BARD)}\\

488752 & organism-based & cell-based & Quantitative high throughput screen for delayed death inhibitors of the malarial parasite plastid, 48 hour incubation & cell-based & as for assay 488745 & Expert agrees with LLM majority: \textcolor{softgreen}{ChEMBL label incorrect (LLMs agree with BARD)}\\

488788 & biochemical & cell-free & Ras-converting Enzyme/Cell Proliferation Pathway Measured in Biochemical System Using Plate Reader & biochemical & a bit ambiguous in the title, but likely biochemical using proteases from either yeast or human & Expert agrees with ChEMBL: \textcolor{softorange}{After seeing evidence from the LLMs: they pointed out: it does say that they do membrane preparations (I was mostly reading the sentence ``This assay has been used for characterization of both yeast and human CaaX proteases and for characterization of inhibitors.''; I tried to look up the reference, but could only see the abstract), so also LLM might be correct, ambiguous case}\\

492957 & cell-based & organism-based & Counterscreen for AddAB inhibitors: absorbance-based bacterial cell-based high throughput dose response assay to identify inhibitors of RecBCD & organism-based & see short title (again, could be defined as organism based \ldots) & Expert agrees with LLM majority: \textcolor{softgreen}{ChEMBL/BARD label incorrect} (expert initially leaned towards ChEMBL, but adapted after discussion) \\

504494 & biochemical & cell-free & Late stage assay provider results from the probe development effort to identify inhibitors of PAFAH2: Fluorescence-based biochemical gel-based Activity-Based Protein Profiling (ABPP) IC50 & biochemical & see short title & Expert agrees with ChEMBL: \textcolor{softorange}{After seeing evidence: ``The protocol definitely says that it uses `soluble proteome' of cells, which your last LLM explicitly defined as `cell lysate', which then would fit in the `cell-free format'. If `biochemical format' requires a purified single protein or other macromolecule, one can also argue that a proteome is not just a single protein, but a mixture and since it is derived from a cell (but not a cell anymore), cell-free is fine. `The assay uses a soluble proteome (cell lysate) of BW5147-derived murine T cells, which is explicitly a cell-free system.' So the question really is whether the `proteome' is purified enough to become `biochemical' -- or not \ldots'', so also LLM might be correct}\\

463208 & biochemical & cell-free & Fluorescence Biochemical Dose Confirmation to Identify Inhibitors of Yeast HDACs & cell-free & Yeast nuclear extracts (YNE) were used; could be biochemical, but does not seem to be an isolated biological macromolecule & Expert agrees with LLM majority: \textcolor{softgreen}{ChEMBL/BARD label incorrect} \\

2096 & cell-based & organism-based & Luminescence Cell-Based Dose Confirmation HTS to Identify Inhibitors of Polyadenylation. & organism-based & Title/protocol (``cells are lysed\ldots'') suggest cell-based, but it is yeast, so probably organism-based & Expert agrees with LLM majority: \textcolor{softgreen}{ChEMBL/BARD label incorrect} \\

2467 & cell-based & organism-based & Fluorescence Cell-Based Retest of C. albicans Growth in the Presence of Fluconazole & organism-based & ``Whole cell phenotypic screening'' would indicate cell-based, however C. albicans is an organism & Expert agrees with LLM majority: \textcolor{softgreen}{ChEMBL label incorrect (LLMs agree with BARD)} \\

2695 & cell-based & organism-based & Fluorescence Cell-Based Dose Screen to Determine Inhibitors of C. albicans Viability & organism-based & already in short title, but later adapted when we discussed definition & Expert agrees with LLM majority: \textcolor{softgreen}{ChEMBL/BARD label incorrect} (expert initially leaned towards ChEMBL, but adapted after discussion) \\

492959 & cell-based & organism-based & Absorbance-based bacterial cell-based high throughput dose response assay for inhibitors of AddAB recombination protein complex & organism-based & see short title (again, could be defined as organism based \ldots), but adapted when discussing & Expert agrees with LLM majority: \textcolor{softgreen}{ChEMBL/BARD label incorrect} (expert initially leaned towards ChEMBL, but adapted after discussion) \\

504631 & organism-based & cell-based & A cell-based HTS for delayed death inhibitors of the malarial parasite plastid Measured in Microorganism System Using Plate Reader & organism-based & Title would indicate cell-based, however Plasmodium falciparum is an organism & Expert agrees with ChEMBL: \textcolor{softred}{LLM majority incorrect (see Appendix~\ref{app:failure_mode_organism})} \\

651893 & cell-based & organism-based & Inhibition of Trypanosoma brucei rhodesiense STIB900 Measured in Cell-Based System Using Plate Reader & organism-based & see short title, adapted when discussing definition & Expert agrees with LLM majority: \textcolor{softgreen}{ChEMBL/BARD label incorrect} (expert initially leaned towards ChEMBL, but adapted after discussion) \\

\end{longtable}
\end{landscape}

\subsection{Failure case: AID 504631}
\label{app:failure_mode_organism}

A notable disagreement concerned assay AID 504631 using the single-celled pathogens \textit{Plasmodium falciparum}. Under the BAO ontology, \textit{cell-based format} is explicitly restricted to eukaryotic cells \textit{derived from a multicellular organism} (cell lines, primary cells, stem cells) and is explicitly distinguished from \textit{organism-based format}, which covers any living organism, single- or multi-celled. Bacteria and yeast are given as the canonical single-celled examples. A single-celled eukaryotic parasite such as \textit{P.\ falciparum} therefore falls under \textit{organism-based}, not \textit{cell-based}, even though it is eukaryotic and is cultured in wells.

Table~\ref{tab:organism_failure} shows model predictions for AID~504631, ``A cell-based HTS for delayed death inhibitors of the malarial parasite''. Four of seven models (Gemma 3 27B, GPT-OSS 20B, GPT-4o, Llama~3.3~70B) predicted \textit{cell-based format} with high confidence, while three (Claude, Gemini, Gemma~4~31B) correctly predicted \textit{organism-based format}. Notably, the erroneous predictions are not due to missing information: the cited evidence in several cases (e.g., GPT-OSS 20B, Llama~3.3~70B) directly quotes text referring to ``the malarial parasite'' or a ``microorganism system,'' yet the models still defaulted to \textit{cell-based}. This suggests the failure is not one of information retrieval but of ontological reasoning: models appear to pattern-match on surface cues such as ``cells cultured in wells'' or the word ``parasite'' co-occurring with ``cell,'' rather than applying the BAO-specific distinction. The assay's own title (``cell-based HTS'') likely reinforces this bias, as several models' cited evidence draws directly on title wording rather than the underlying biology.

\begin{table}[h]
\centering
\small
\caption{Model predictions for AID~504631, illustrating the cell-based/organism-based confusion for single-celled eukaryotic organisms. True label: \textit{organism-based format}.}
\label{tab:organism_failure}
\begin{tabular}{@{}l l l p{7.5cm}@{}}
\toprule
\textbf{Model} & \textbf{Prediction} & \textbf{Correct?} & \textbf{Cited evidence} \\
\midrule
Claude & organism-based & \checkmark & ``...parasites cultured in the presence of test compounds...measuring susceptibility of the living malarial parasite organism.'' \\
Gemini & organism-based & \checkmark & ``The susceptibility of the Dd2 \textit{Plasmodium falciparum} line against novel small molecules will be determined using a SYBR green-based fluorescence assay.'' \\
Gemma~4~31B & organism-based & \checkmark & ``The susceptibility of the Dd2 \textit{Plasmodium falciparum} line...Parasites are cultured in the presence of serial dilutions of test compounds.'' \\
Gemma 3 27B & cell-based & \ding{55} & ``A cell-based HTS for delayed death inhibitors of the malarial parasite plastid...Parasites are cultured in the presence of serial dilutions of test compounds...'' \\
GPT-OSS 20B & cell-based & \ding{55} & ``Parasites are cultured in the presence of serial dilutions of test compounds...using a SYBR green-based fluorescence assay in 384-well plates'' \\
GPT-4o & cell-based & \ding{55} & ``...explicitly indicates the use of living cells (parasites) in the experimental setup.'' \\
Llama~3.3~70B & cell-based & \ding{55} & ``A cell-based HTS for delayed death inhibitors of the malarial parasite plastid Measured in Microorganism System Using Plate Reader'' \\
\bottomrule
\end{tabular}
\end{table}
\clearpage
\section{Detection Disagreement Details}
\label{app:full_disagreement_table_detection}

This appendix reports the full record-level breakdown of the 34
physical detection method assays where the majority of LLMs disagreed with
the PubChem bulk-download label
(Table~\ref{tab:appendix_detection_disagreement}), followed by the
per-model evidence spans for which the expert did not agree with the majority of LLMs.
(Table~\ref{tab:detection_predictions}). 

\begingroup
\small
\setlength{\tabcolsep}{3pt}
\renewcommand{\arraystretch}{1.0}

\begin{longtable}{@{}r L{2.3cm} L{2.3cm} L{3.4cm} L{2.8cm} r@{}}
\caption{Manual inspection of the 34 physical detection method assays where the majority of models
disagreed with the third-party BAO detection label in the PubChem bulk download. All labels are BAO
physical detection methods; the ``\ method'' suffix is omitted for space. $n_\text{maj.}$ is the number
of models supporting the majority prediction; $n_\text{al.}$ is the number of predictions that aligned
with the third-party BAO label instead. Seven models were attempted per row unless noted otherwise.
Rows shaded in gray were additionally reviewed by a senior industrial curator
(see Table~\ref{tab:Expert_Review_Detection} for per-model evidence spans).}
\label{tab:appendix_detection_disagreement} \\
\toprule
\textbf{AID} & \textbf{BAO label (third-party)} & \textbf{Majority pred.} & \textbf{Bioassay annotation (PubChem)} & \textbf{Comment} & \textbf{$n_\text{maj.}$/$n_\text{al.}$} \\
\midrule
\endfirsthead

\multicolumn{6}{c}{\tablename\ \thetable\ -- \textit{continued from previous page}} \\
\toprule
\textbf{AID} & \textbf{BAO label (third-party)} & \textbf{Majority pred.} & \textbf{Bioassay annotation (PubChem)} & \textbf{Comment} & \textbf{$n_\text{maj.}$/$n_\text{al.}$} \\
\midrule
\endhead

\midrule
\multicolumn{6}{r}{\textit{continued on next page}} \\
\endfoot

\bottomrule
\endlastfoot

588725  & spectrophotometry & radiometry & scintillation counting (part of radiometry) & LLMs agree with BARD & 7/0 \\
588768  & fluorescence & spectrophotometry & absorbance (part of spectrophotometry) & LLMs agree with BARD & 7/0 \\
588782  & fluorescence & luminescence & luminescence & LLMs agree with BARD & 7/0 \\
602192  & fluorescence & luminescence & chemiluminescence (part of luminescence) & LLMs agree with BARD & 7/0 \\
602194  & fluorescence & luminescence & chemiluminescence (part of luminescence) & LLMs agree with BARD & 7/0 \\
602231  & spectrophotometry & isometric tension recording & label-free method (BARD) & LLMs disagree with all sources & 7/0 \\
602245  & spectrophotometry & isometric tension recording & (no other) & -- & 7/0 \\
602337  & fluorescence & spectrophotometry & absorbance (part of spectrophotometry) & LLMs agree with BARD & 7/0 \\
602339  & fluorescence & mass spectrometry & mass spectrometry & LLMs agree with BARD & 7/0 \\
602341  & fluorescence & mass spectrometry & mass spectrometry & LLMs agree with BARD & 7/0 \\
602351  & fluorescence & mass spectrometry & mass spectrometry & LLMs agree with BARD & 7/0 \\
602353  & fluorescence & mass spectrometry & mass spectrometry & LLMs agree with BARD & 7/0 \\
624079  & spectrophotometry & fluorescence & fluorescence & LLMs agree with BARD & 7/0 \\
1259354  & fluorescence & luminescence & (no other) & LLM prediction seems reasonable & 6/0 \\
\rowcolor{gray!15}
623968  & spectrophotometry & fluorescence method & quantitative reverse transcription PCR (this is not a physical detection method but an assay supporting method) & -- & 3/3 (6 att.) \\
\rowcolor{gray!15}
588778  & fluorescence & spectrophotometry & (no other) & -- & 5/2 \\
\rowcolor{gray!15}
602418  & luminescence & fluorescence & (no other) & -- & 3/0 \\
\rowcolor{gray!15}
588816  & fluorescence & simple measurement method & (no other) & -- & 3/1 (6 att.) \\
\rowcolor{gray!15}
602170  & spectrophotometry & visual observation method & (no other) & -- & 3/1 \\
\rowcolor{gray!15}
624103  & fluorescence & simple measurement method & (no other) & -- & 3/2 (6 att.) \\
\rowcolor{gray!15}
624135  & fluorescence & simple measurement method & (no other) & -- & 3/2 (6 att.) \\
\rowcolor{gray!15}
624334  & fluorescence & visual observation method & (no other) & -- & 2/0 (6 att.) \\
623916  & spectrophotometry & simple measurement & cell number determination (not a BAO format, but assay design method) & -- & 5/0 \\
623917  & radiometry & simple measurement & cell number determination (not a BAO format, but assay design method) & -- & 5/0 \\
\rowcolor{gray!15}
623918  & spectrophotometry & simple measurement method & cell number determination (not a BAO format, but assay design method) & -- & 4/0 \\
623919  & spectrophotometry & simple measurement & cell number determination (not a BAO format, but assay design method) & -- & 5/0 \\
623930  & radiometry & luminescence & luminescence & LLMs agree with BARD & 6/0 (6 att.) \\
623954  & spectrophotometry & simple measurement & cell number determination & -- & 4/0 \\
623956  & spectrophotometry & simple measurement & cell number determination & -- & 4/0 \\
624085  & spectrophotometry & fluorescence & fluorescence & LLMs agree with BARD & 6/1 \\
624090  & spectrophotometry & fluorescence & fluorescence & LLMs agree with BARD & 6/1 \\
624092  & spectrophotometry & fluorescence & fluorescence & LLMs agree with BARD & 6/1 \\
\rowcolor{gray!15}
624286  & luminescence & visual observation & plaque counting (not BAO label) & -- & 6/0 (6 att.) \\
\rowcolor{gray!15}
651578  & luminescence & visual observation & plaque counting (not BAO label) & -- & 6/0 (6 att.) \\
\end{longtable}
\endgroup

\begingroup
\small
\setlength{\tabcolsep}{4pt}
\renewcommand{\arraystretch}{1.15}

\begin{longtable}{@{}r L{2.2cm} L{2.2cm} L{3.2cm} L{2.2cm} L{3.5cm} @{}}
\caption{Detection-method assays sent for expert review, with third-party (PubChem) label, LLM majority prediction, and expert annotation.}
\label{tab:Expert_Review_Detection} \\

\toprule
\textbf{AID} & \textbf{3rd-party} & \textbf{Majority pred.} & \textbf{Short title} & \textbf{Expert format} & \textbf{Expert reason} \\
\midrule
\endfirsthead

\multicolumn{6}{c}{\tablename\ \thetable\ -- \textit{continued from previous page}} \\
\toprule
\textbf{AID} & \textbf{3rd-party} & \textbf{Majority pred.} & \textbf{Short title} & \textbf{Expert format} & \textbf{Expert reason}\\
\midrule
\endhead

\midrule
\multicolumn{6}{r}{\textit{continued on next page}} \\
\endfoot

\bottomrule
\endlastfoot

623968  & spectrophotometry & fluorescence & Inhibitors of the Hepatitis C Virus non-structural protein 3 helicase (NS3) & spectrophotometry & Similar work to the CHRM probe-development assays but ``late stage,'' so more information is given, e.g.\ ``RNA concentration is determined by reading absorbance at 260\,nm''  \\

588778  & fluorescence & spectrophotometry & A High Throughput Screening Assay for Inhibitors of Bacterial Motility in \textit{Vibrio cholerae} & fluorescence (spectroscopy also fine after seeing LLM evidence) & Protocol: ``\ldots fluorescence intensity was determined by reading\ldots''\\

602418  & luminescence & fluorescence & Summary of probe development efforts to identify inverse agonists of LRH-1 (NR5A2) & cannot define & A summary of various efforts per the title; no information on what was actually done for each effort  \\

588816  & fluorescence & simple measurement & Identify agonists of the human cholinergic receptor, muscarinic 1 (CHRM1) & cannot be defined & No information; this is pre-work to identify a probe, not really an assay in itself  \\

602170  & spectrophotometry & visual observation & Probe development efforts to identify activators of Methionine sulfoxide reductase A (MsrA) & cannot be defined & As above (probe-development pre-work)  \\

624103  & fluorescence & simple measurement & Identify agonists of the human cholinergic receptor, muscarinic 5 (CHRM5) & cannot be defined & As above \\

624135  & fluorescence & simple measurement & Identify agonists, PAMs, and antagonists of the human cholinergic receptor, muscarinic 4 (CHRM4) & cannot be defined & As above  \\

624334  & fluorescence & visual observation & Small molecule inhibitors of the oncogenic and cytokinetic protein MgcRacGAP & cannot be defined & No information \\

623918  & spectrophotometry & simple measurement & Cell-based colony formation assay to identify inhibitors of the recombination-promoting activity of RecBCD in V66 \textit{E.\ coli} & simple measurement & Not exactly the same assay as 623916/623917 but same principle  \\

624286  & luminescence & visual observation & Virus Titer Reduction Secondary Screen for Compounds that Inhibit VEEV (TC-83 strain) & not easy -- simple measurement, visual observation, or imaging & Crystal violet solution added; number of plaques per well determined by visual counting; expert reluctant to use ``visual observation,'' reserving that category for larger-organism behavioral readouts \\

651578  & luminescence & visual observation & As above (VEEV TC-83 titer reduction) & as above (624286) & As above \\

\end{longtable}
\endgroup

\begingroup
\footnotesize
\setlength{\tabcolsep}{4pt}
\renewcommand{\arraystretch}{1.05}

\begin{longtable}{@{}p{0.8cm} p{2.5cm} p{2.5cm} p{1.7cm} p{2.7cm} p{1.7cm} p{3.3cm}@{}}
\caption{Detection technology predictions for selected AIDs across all models.}
\label{tab:detection_predictions} \\
\toprule
\textbf{AID} & \textbf{3rd-Party Label} & \textbf{Expert Label} & \textbf{Model} & \textbf{Prediction} & \textbf{Conf.} & \textbf{Evidence Span} \\
\midrule
\endfirsthead
\multicolumn{7}{c}{\tablename\ \thetable{} -- continued} \\
\toprule
\textbf{AID} & \textbf{3rd-Party Label} & \textbf{Expert Label} & \textbf{Model} & \textbf{Prediction} & \textbf{Conf.} & \textbf{Evidence Span} \\
\midrule
\endhead
\midrule \multicolumn{7}{r}{\textit{Continued on next page}} \\
\endfoot
\bottomrule
\endlastfoot
\multirow{7}{*}{588816} & \multirow{7}{*}{\parbox{1.5cm}{fluorescence method}} & \multirow{7}{*}{\parbox{1.5cm}{cannot be defined}} & Llama 3.3 70B & simple measurement \ding{55} & low & No specific detection method mentioned \\
 &  &  & GPT-4o & visual observation \ding{55} & low & No specific detection method or instrumentation mentioned. \\
 &  &  & Claude Sonnet 4.6 & ERROR \ding{55} &  &  \\
 &  &  & Gemini 3.6 Flash & fluorescence \ding{55} & low & Title indicates probe development for human muscarinic receptor 1 (CHRM1), typically fluorescence-based\ldots{} \\
 &  &  & Gemma 3 27B & simple measurement \ding{55} & low & No information provided. Conservative assumption\ldots{} \\
 &  &  & Gemma 4 31B & label free \ding{55} & low & probe development efforts to identify agonists of CHRM1 \\
 &  &  & GPT-OSS 20B & simple measurement \ding{55} & low & No detection method described \\

\multirow{7}{*}{602170} & \multirow{7}{*}{\parbox{1.5cm}{spectro- photometry method}} & \multirow{7}{*}{\parbox{1.5cm}{cannot be defined}} & Llama 3.3 70B & fluorescence \ding{55} & low & No explicit evidence, but MsrA assays often associated with fluorescence-based detection\ldots{} \\
 &  &  & GPT-4o & visual observation \ding{55} & low & No explicit detection method or instrumentation mentioned. \\
 &  &  & Claude Sonnet 4.6 & spectrophotometry \ding{55} & low & MsrA activity assays commonly use spectrophotometric or fluorescence-based detection\ldots{} \\
 &  &  & Gemini 3.6 Flash & fluorescence \ding{55} & low & Inferred from typical probe development assays for MsrA \\
 &  &  & Gemma 3 27B & simple measurement \ding{55} & low & No information provided; defaulting to most basic method\ldots{} \\
 &  &  & Gemma 4 31B & visual observation \ding{55} & low & Title mentions 'probe development efforts', often visual/basic observation\ldots{} \\
 &  &  & GPT-OSS 20B & visual observation \ding{55} & low & No protocol provided \\
\midrule
\multirow{7}{*}{623918} & \multirow{7}{*}{\parbox{1.5cm}{spectro- photometry method}} & \multirow{7}{*}{\parbox{1.5cm}{simple measurement}} & Llama 3.3 70B & simple measurement \checkmark & high & colony forming units per ml of culture \\
 &  &  & GPT-4o & visual observation \ding{55} & high & Differential plating to measure colony formation relies on visual observation\ldots{} \\
 &  &  & Claude Sonnet 4.6 & simple measurement \checkmark & medium & Frequency of recombinants determined by differential plating and counting\ldots{} \\
 &  &  & Gemini 3.6 Flash & visual observation \ding{55} & high & Frequency of recombinants determined by differential plating and calculating\ldots{} \\
 &  &  & Gemma 3 27B & simple measurement \checkmark & high & Counting colonies (colony forming units) to determine viability\ldots{} \\
 &  &  & Gemma 4 31B & visual observation \ding{55} & high & Frequency of recombinants determined by differential plating \\
 &  &  & GPT-OSS 20B & simple measurement \checkmark & high & Plated to measure viability and determine recombinant frequency \\
\midrule
\multirow{7}{*}{624103} & \multirow{7}{*}{\parbox{1.5cm}{fluorescence method}} & \multirow{7}{*}{\parbox{1.5cm}{cannot be defined}} & Llama 3.3 70B & simple measurement \ding{55} & low & No specific detection method mentioned \\
 &  &  & GPT-4o & visual observation \ding{55} & low & No explicit detection method, techniques, or instrumentation mentioned\ldots{} \\
 &  &  & Claude Sonnet 4.6 & fluorescence \ding{55} & low & Muscarinic receptor agonist assays commonly use fluorescence-based detection\ldots{} \\
 &  &  & Gemini 3.6 Flash & fluorescence \ding{55} & low & No explicit physical detection method described. \\
 &  &  & Gemma 3 27B & simple measurement \ding{55} & low & No detection method stated; lack of detail\ldots{} \\
 &  &  & Gemma 4 31B & ERROR \ding{55} &  &  \\
 &  &  & GPT-OSS 20B & simple measurement \ding{55} & low & No protocol or description provided \\
\midrule
\multirow{7}{*}{624135} & \multirow{7}{*}{\parbox{1.5cm}{fluorescence method}} & \multirow{7}{*}{\parbox{1.5cm}{cannot be defined}} & Llama 3.3 70B & simple measurement \ding{55} & low & No specific detection method mentioned \\
 &  &  & GPT-4o & visual observation \ding{55} & low & No explicit detection method, techniques, or instrumentation mentioned\ldots{} \\
 &  &  & Claude Sonnet 4.6 & fluorescence \ding{55} & low & Muscarinic receptor assays commonly use fluorescence-based detection (e.g., HTRF)\ldots{} \\
 &  &  & Gemini 3.6 Flash & fluorescence \ding{55} & low & CHRM4 probe development typically utilizes fluorometric functional assays\ldots{} \\
 &  &  & Gemma 3 27B & simple measurement \ding{55} & low & Only indicates probe development efforts, no detection method\ldots{} \\
 &  &  & Gemma 4 31B & ERROR \ding{55} &  &  \\
 &  &  & GPT-OSS 20B & simple measurement \ding{55} & low & No detection method described \\
\midrule
\multirow{7}{*}{624334} & \multirow{7}{*}{\parbox{1.5cm}{fluorescence method}} & \multirow{7}{*}{\parbox{1.5cm}{cannot be defined}} & Llama 3.3 70B & simple measurement \ding{55} & low & No specific detection method mentioned \\
 &  &  & GPT-4o & visual observation \ding{55} & low & No explicit detection method or instrumentation mentioned\ldots{} \\
 &  &  & Claude Sonnet 4.6 & spectrophotometry \ding{55} & low & Classification based on common methods used in small molecule inhibitor screens\ldots{} \\
 &  &  & Gemini 3.6 Flash & luminescence \ding{55} & low & No explicit method mentioned; luminescence inferred as common\ldots{} \\
 &  &  & Gemma 3 27B & simple measurement \ding{55} & low & No information provided; defaulting to most basic\ldots{} \\
 &  &  & Gemma 4 31B & ERROR \ding{55} &  &  \\
 &  &  & GPT-OSS 20B & visual observation \ding{55} & low & No detection method described \\
\midrule
\multirow{7}{*}{623968} & \multirow{7}{*}{\parbox{1.5cm}{spectro- photometry method}} & \multirow{7}{*}{\parbox{1.5cm}{spectro- photometry method}} & Llama 3.3 70B & spectrophotometry \checkmark & high & RNA concentration determined by reading absorbance at 260 nm \\
 &  &  & GPT-4o & spectrophotometry \checkmark & high & RNA concentration determined by reading absorbance at 260 nm. \\
 &  &  & Claude Sonnet 4.6 & ERROR \ding{55} &  &  \\
 &  &  & Gemini 3.6 Flash & fluorescence \ding{55} & high & Real time RT-PCR performed with TaqMan chemistry \\
 &  &  & Gemma 3 27B & spectrophotometry \checkmark & high & RNA concentration determined by reading absorbance at 260 nm. \\
 &  &  & Gemma 4 31B & fluorescence \ding{55} & high & RT-PCR using TaqMan chemistry used to quantify HCV RNA levels \\
 &  &  & GPT-OSS 20B & fluorescence \ding{55} & high & Real time RT-PCR performed with TaqMan chemistry \\
\midrule
\multirow{7}{*}{588778} & \multirow{7}{*}{\parbox{1.5cm}{fluorescence method}} & \multirow{7}{*}{\parbox{1.5cm}{fluorescence \& spectrophotometry}} & Llama 3.3 70B & spectrophotometry \checkmark & high & Absorbance at 615nm on a Perkin Elmer Envision \\
 &  &  & GPT-4o & fluorescence \checkmark & high & Fluorescence intensity determined using excitation/emission to differentiate motility\ldots{} \\
 &  &  & Claude Sonnet 4.6 & spectrophotometry \checkmark & high & Primary read was absorbance at 615nm to monitor bacterial growth and motility\ldots{} \\
 &  &  & Gemini 3.6 Flash & spectrophotometry \checkmark & high & Primary read was absorbance at 615 to monitor bacterial growth and motility\ldots{} \\
 &  &  & Gemma 3 27B & fluorescence \checkmark & high & Fluorescence intensity determined by reading with excitation wavelength of 535\ldots{} \\
 &  &  & Gemma 4 31B & spectrophotometry \checkmark & high & Read using absorbance at 615nm on a Perkin Elmer Envision\ldots{} \\
 &  &  & GPT-OSS 20B & spectrophotometry \checkmark & high & "read using absorbance at 615nm on a Perkin Elmer Envision" \\
\midrule
\multirow{7}{*}{602418} & \multirow{7}{*}{\parbox{1.5cm}{luminescence method}} & \multirow{7}{*}{\parbox{1.5cm}{cannot define}} & Llama 3.3 70B & fluorescence \ding{55} & low & No explicit evidence, but fluorescence commonly used in HTS for inverse agonists \\
 &  &  & GPT-4o & visual observation \ding{55} & low & No explicit detection method or instrumentation mentioned. \\
 &  &  & Claude Sonnet 4.6 & fluorescence \ding{55} & low & LRH-1 probe development assays commonly use fluorescence-based methods (e.g., TR-FRET)\ldots{} \\
 &  &  & Gemini 3.6 Flash & fluorescence \ding{55} & low & Probe development for LRH-1 typically utilizes TR-FRET/fluorescence assays\ldots{} \\
 &  &  & Gemma 3 27B & simple measurement \ding{55} & low & No information provided; assuming basic assay\ldots{} \\
 &  &  & Gemma 4 31B & label free \ding{55} & low & Probe development to identify inverse agonists often uses label-free methods like SPR\ldots{} \\
 &  &  & GPT-OSS 20B & simple measurement \ding{55} & low & No detection method described \\
\midrule
\multirow{7}{*}{624286} & \multirow{7}{*}{\parbox{1.5cm}{luminescence method}} & \multirow{7}{*}{\parbox{1.5cm}{simple meas., visual obs., or imaging}} & Llama 3.3 70B & visual observation \checkmark & high & Number of plaques in each well determined by visual counting. \\
 &  &  & GPT-4o & visual observation \checkmark & high & Protocol states number of plaques determined by visual counting\ldots{} \\
 &  &  & Claude Sonnet 4.6 & ERROR \ding{55} &  &  \\
 &  &  & Gemini 3.6 Flash & visual observation \checkmark & high & Number of plaques in each well determined by visual counting \\
 &  &  & Gemma 3 27B & visual observation \checkmark & high & Relies on 'visual counting' of plaques after crystal violet staining\ldots{} \\
 &  &  & Gemma 4 31B & visual observation \checkmark & high & Number of plaques in each well determined by visual counting. \\
 &  &  & GPT-OSS 20B & visual observation \checkmark & high & "number of plaques in each well determined by visual counting" \\
\midrule
\multirow{7}{*}{651578} & \multirow{7}{*}{\parbox{1.5cm}{luminescence method}} & \multirow{7}{*}{\parbox{1.5cm}{not easy; possibly visual}} & Llama 3.3 70B & visual observation \checkmark & high & Number of plaques in each well determined by visual counting. \\
 &  &  & GPT-4o & visual observation \checkmark & high & Protocol states number of plaques determined by visual counting\ldots{} \\
 &  &  & Claude Sonnet 4.6 & ERROR \ding{55} &  &  \\
 &  &  & Gemini 3.6 Flash & visual observation \checkmark & high & Number of plaques in each well determined by visual counting \\
 &  &  & Gemma 3 27B & visual observation \checkmark & high & Relies on 'visual counting' of plaques after crystal violet staining\ldots{} \\
 &  &  & Gemma 4 31B & visual observation \checkmark & high & Plates incubated and stained again; washed after\ldots{} \\
 &  &  & GPT-OSS 20B & visual observation \checkmark & high & "number of plaques in each well determined by visual counting" \\

\end{longtable}
\endgroup

\clearpage

\section{Confidence calibration on both tasks}
\label{app:confidence_calibration}
\begin{figure}[htbp]
    \centering
    
    \begin{subfigure}[b]{\linewidth}
        \centering
        \includegraphics[width=\linewidth]{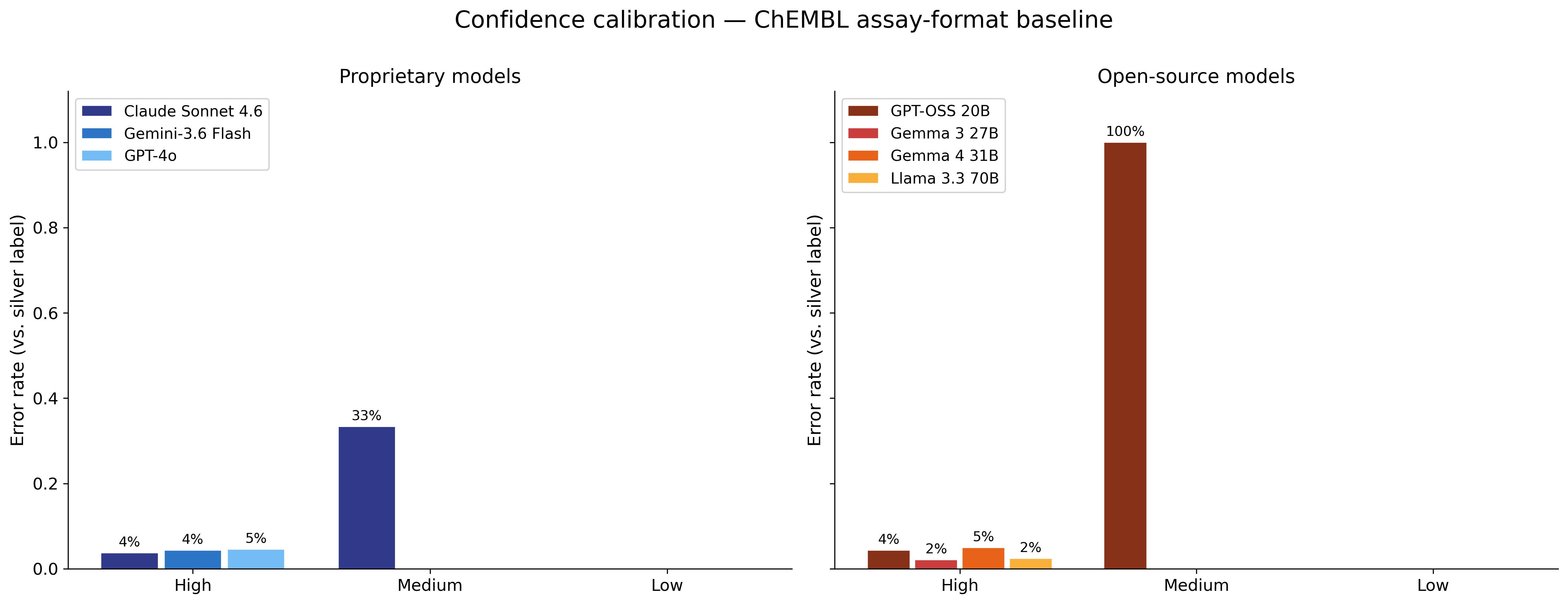}
        \caption{Error rate by predicted confidence level for silver-label classification in the ChEMBL baseline dataset.}
        \label{fig:confidence_ChEMBL}
    \end{subfigure}
    \hfill
    \begin{subfigure}[b]{\linewidth}
        \centering
        \includegraphics[width=\linewidth]{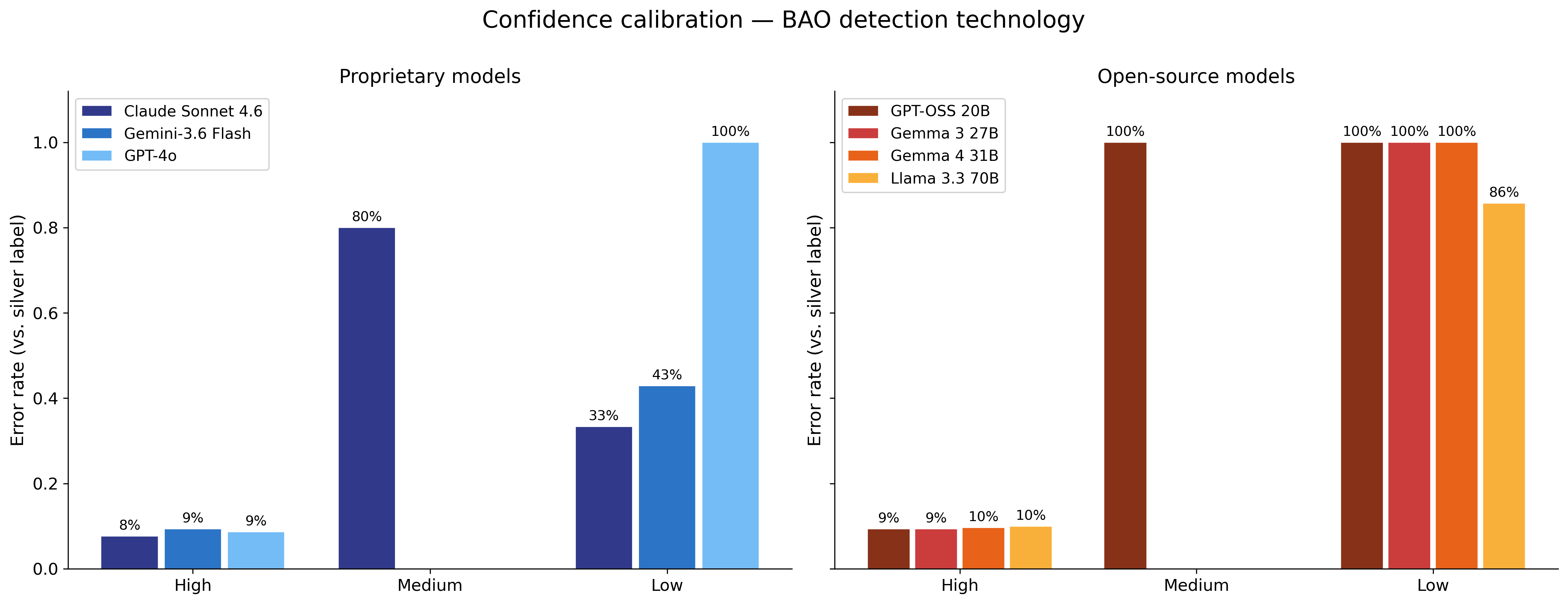}
        \caption{Error rate by predicted confidence level for silver-label detection results.}
        \label{fig:detection_confidence_levels}
    \end{subfigure}
    
    \caption{Error rates across predicted confidence levels for silver-label predictions. The top panel shows results for the ChEMBL baseline, while the bottom panel shows the corresponding results for the detection task. Each bar represents the proportion of incorrect predictions within a confidence category, with the number of samples shown in parentheses.}
    \label{fig:confidence_comparison}
\end{figure}

\clearpage

\section{Prompt ablation details}
\label{app:prompt_ablations}

\begin{figure}[htbp]
    \centering
    \includegraphics[width=\linewidth]{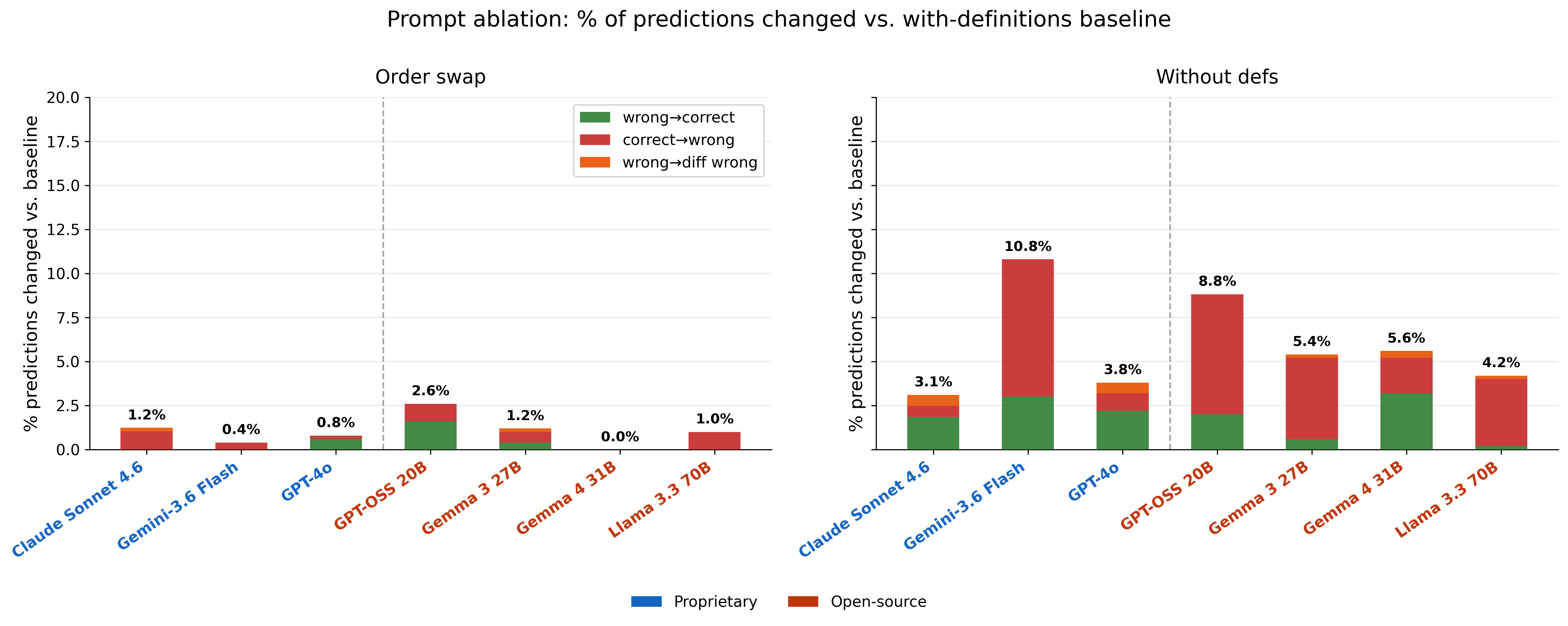}
    \caption{Effect of two prompt perturbations on proprietary-LLM
    predictions for the ChEMBL assay format subset. Removing the BAO
    definitions (Without defs) changes far more predictions
    than reversing the class order (Order swap), and
    disproportionately so for Gemini 3.6 Flash (10.8\%).
    Bar segments indicate whether each changed prediction moved from a different label to the silver label (green), from the silver label to another label (red), or between two
    labels that are different from the silver label (orange); percentages above each bar give the
    total fraction of predictions that changed.}
    \label{fig:Changed_labels_prompt_abblations}
\end{figure}
\clearpage

\section{Resource Usage: Cost and Energy}
\label{app: resource usage}

This appendix reports the compute cost, wall-clock throughput, and (proxy)
energy consumption of the LLM auditing workflow.

The open-weights models were served locally via \texttt{Ollama v0.17.4}
on a single node of the Berzelius cluster. Wall-clock time was
measured per SLURM job. Note that this could be improved for open-source models. The energy per record is a proxy computed as
$\text{GPU-hours} \times P_\text{GPU}$. We treat
this as a proxy: it excludes non-GPU consumption, such as CPU and cooling overhead. For rerunning with the same settings we would like to declare that GPT-OSS 20B and Gemma 3 27B and Gemma 4 were hosted on a single NVIDIA A100-SXM4-40GB, LLaMA 3.3 70B on a single NVIDIA H200, each with 32 CPU cores allocated.

Table~\ref{tab:compute_accounting} reports per-model resource usage across all five experimental conditions. Proprietary models (Claude Sonnet~4.6, Gemini~3.6~Flash, GPT-4o) were accessed via API and are billed per token; we report USD cost per completion. Open-weight models (Gemma 3 27B~27B, Gemma 4~31B, GPT-OSS~20B, Llama~3.3~70B) were self-hosted, incurring no API cost but measurable energy draw; we report energy consumption per completion in place of a dollar cost. No energy figures are available for the proprietary models, as token-level energy accounting is not exposed by the respective providers. 

\begingroup
\small
\setlength{\tabcolsep}{4pt}
\renewcommand{\arraystretch}{1}
\begin{longtable}{@{}l l r r r r r r@{}}
\caption{Per-model token usage, parse failures, and cost/energy accounting across all experimental conditions.
Input and completion token counts are experiment-wide totals.
Cost is reported for API-accessed (proprietary) models; energy for self-hosted (open-weight) models.
``--'' indicates not applicable.}
\label{tab:compute_accounting} \\
\toprule
\textbf{Experiment} & \textbf{Model} & \textbf{$N$} & \textbf{Parse} & \textbf{Input} & \textbf{Output} & \textbf{Cost} & \textbf{Energy} \\
                    &                &              & \textbf{fails} & \textbf{tokens} & \textbf{tokens} & \textbf{(\$)} & \textbf{(kWh)} \\
\midrule
\endfirsthead

\multicolumn{8}{c}{\tablename\ \thetable\ -- \textit{continued from previous page}} \\
\toprule
\textbf{Experiment} & \textbf{Model} & \textbf{$N$} & \textbf{Parse} & \textbf{Input} & \textbf{Output} & \textbf{Cost} & \textbf{Energy} \\
                    &                &              & \textbf{fails} & \textbf{tokens} & \textbf{tokens} & \textbf{(\$)} & \textbf{(kWh)} \\
\midrule
\endhead

\midrule
\multicolumn{8}{r}{\textit{continued on next page}} \\
\endfoot

\bottomrule
\endlastfoot

\multirow{7}{*}{ChEMBL baseline}
 & Claude Sonnet~4.6 & 1097 & \textbf{29} & 2{,}302{,}201 & 90{,}590 & 9.09 & --   \\
 & Gemini~3.6~Flash  & 1097 &  0           & 2{,}040{,}815 & 54{,}349 & 3.47 & --   \\
 & GPT-4o            & 1097 &  0           & 1{,}998{,}562 & 67{,}554 & 5.67 & --   \\
 & Gemma~3~27B       & 1097 &  0           & 2{,}081{,}404 & 64{,}558 & --   & 6.44 \\
 & Gemma~4~31B       & 1097 &  0           & 2{,}086{,}917 & 63{,}083 & --   & 6.45 \\
 & GPT-OSS~20B       & 1097 &  0           & 2{,}236{,}521 & 51{,}689 & --   & 6.85 \\
 & Llama~3.3~70B     & 1097 &  0           & 2{,}008{,}532 & 52{,}330 & --   & 6.18 \\
\midrule

\multirow{7}{*}{Order swap}
 & Claude Sonnet~4.6 & 500 & \textbf{12} & 1{,}053{,}952 & 41{,}224 & 4.16 & --   \\
 & Gemini~3.6~Flash  & 500 &  0           &   939{,}568   & 24{,}503 & 1.59 & --   \\
 & GPT-4o            & 500 &  0           &   907{,}504   & 30{,}607 & 2.57 & --   \\
 & Gemma~3~27B       & 500 &  0           &   946{,}568   & 29{,}901 & --   & 2.93 \\
 & Gemma~4~31B       & 500 &  0           &   948{,}589   & 30{,}711 & --   & 2.94 \\
 & GPT-OSS~20B       & 500 &  0           & 1{,}013{,}895 & 24{,}958 & --   & 3.12 \\
 & Llama~3.3~70B     & 500 &  0           &   912{,}284   & 24{,}164 & --   & 2.81 \\
\midrule

\multirow{7}{*}{PubChem+defs}
 & Claude Sonnet~4.6 & 226 & \textbf{21} & 417{,}023 & 19{,}332 & 1.70 & --   \\
 & Gemini~3.6~Flash  & 226 &  0           & 363{,}743 & 11{,}523 & 0.63 & --   \\
 & GPT-4o            & 226 &  0           & 362{,}937 & 13{,}537 & 1.04 & --   \\
 & Gemma~3~27B       & 226 &  0           & 372{,}331 & 12{,}763 & --   & 1.16 \\
 & Gemma~4~31B       & 226 &  0           & 373{,}235 & 13{,}367 & --   & 1.16 \\
 & GPT-OSS~20B       & 226 &  0           & 410{,}707 & 10{,}608 & --   & 1.26 \\
 & Llama~3.3~70B     & 226 &  0           & 365{,}753 & 10{,}983 & --   & 1.13 \\
\midrule

\multirow{7}{*}{Without defs}
 & Claude Sonnet~4.6 & 500 & \textbf{13} & 840{,}452 & 39{,}411 & 3.42 & --   \\
 & Gemini~3.6~Flash  & 500 &  0           & 757{,}068 & 22{,}889 & 1.31 & --   \\
 & GPT-4o            & 500 &  0           & 722{,}504 & 29{,}221 & 2.10 & --   \\
 & Gemma~3~27B       & 500 &  0           & 764{,}068 & 29{,}382 & --   & 2.38 \\
 & Gemma~4~31B       & 500 &  0           & 766{,}089 & 27{,}634 & --   & 2.38 \\
 & GPT-OSS~20B       & 500 &  1           & 834{,}022 & 31{,}697 & --   & 2.60 \\
 & Llama~3.3~70B     & 500 &  0           & 726{,}784 & 21{,}688 & --   & 2.25 \\
\midrule

\multirow{7}{*}{Detection}
 & Claude Sonnet~4.6 & 308 & \textbf{34} & 786{,}449 & 21{,}815 & 2.69 & --   \\
 & Gemini~3.6~Flash  & 308 &  0           & 680{,}024 & 15{,}790 & 1.03 & --   \\
 & GPT-4o            & 308 &  0           & 682{,}481 & 17{,}906 & 1.89 & --   \\
 & Gemma~3~27B       & 308 &  0           & 684{,}336 & 18{,}903 & --   & 2.11 \\
 & Gemma~4~31B       & 308 & \textbf{3}  & 685{,}861 & 41{,}668 & --   & 2.18 \\
 & GPT-OSS~20B       & 308 &  0           & 737{,}417 & 13{,}925 & --   & 2.25 \\
 & Llama~3.3~70B     & 308 &  0           & 689{,}615 & 15{,}818 & --   & 2.12 \\

\end{longtable}
\endgroup

\end{document}